\documentclass[runningheads]{llncs}

\usepackage{eccv}

\usepackage{eccvabbrv}

\usepackage{graphicx}
\usepackage{booktabs}

\usepackage[accsupp]{axessibility}

\usepackage{listings}
\usepackage{caption}
\DeclareCaptionType{listing}[Algorithm][List of Algorithms]
\newenvironment{code}{\captionsetup{type=listing}}{}
\usepackage{float}
\usepackage{tcolorbox}

\usepackage{fontawesome5}
\usepackage{animate}
\usepackage{multirow}
\usepackage{soul}         % For \sout in editor macros
\usepackage{wrapfig}      % For text-wrapped figures
\usepackage{nth}

\definecolor{codebg}{RGB}{248,248,248}
\definecolor{codegreen}{rgb}{0.13,0.55,0.13}
\definecolor{codegray}{rgb}{0.5,0.5,0.5}
\definecolor{codepurple}{rgb}{0.58,0,0.82}
\definecolor{codeblue}{rgb}{0.0,0.0,0.7}

\lstnewenvironment{pycode}[1][]{\lstset{#1}}{}

\renewcommand*{\thelisting}{Algorithm \arabic{listing}}

\definecolor{Orange}{HTML}{F4782F}
\definecolor{DeepBlue}{HTML}{0082C9}
\definecolor{MyPink}{HTML}{F1708E}
\definecolor{figure3b}{HTML}{173047}
\definecolor{Pipeline1}{HTML}{EB6C8B}
\definecolor{Pipeline2}{HTML}{012F45}
\definecolor{Pipeline3}{HTML}{EE761C}
\definecolor{yzybest}{rgb}{0.98, 0.8, 0.8}
\definecolor{yzysecond}{rgb}{0.99, 0.88, 0.77}
\definecolor{yzythird}{rgb}{1.0, 1.0, 0.8}

\providecommand{\citet}[1]{\cite{#1}}

\newcommand{\beginsupplement}{%
  \setcounter{table}{0}
  \renewcommand{\thetable}{S\arabic{table}}%
  \setcounter{figure}{0}
  \renewcommand{\thefigure}{S\arabic{figure}}%
  \setcounter{section}{0}
  \renewcommand*{\thelisting}{Algorithm S\arabic{listing}}
}

\newif\ifshowedits
\newcommand{\addeditor}[3]{%
  \definecolor{#1color}{rgb}{#3}
  \expandafter\newcommand\csname #1\endcsname[1]{%
  \ifshowedits
    {\color{#1color} ##1}%
  \else
    {##1}%
  \fi
  }%
  \expandafter\newcommand\csname #1rmk\endcsname[1]{%
  \ifshowedits
    {\color{#1color} {\bf [#2: ##1]}}
  \fi
  }%
  \expandafter\newcommand\csname #1rpl\endcsname[2]{%
  \ifshowedits
    {{\color{#1color} ##1} \sout{##2}}
  \else
    {##1}
  \fi
  }%
}

\usepackage[breaklinks,colorlinks,citecolor=eccvblue]{hyperref}

\usepackage{orcidlink}

\begin{document}

% --- Editor setup ---
\addeditor{junyi}{JW}{0.0, 0.0, 0.8}
\addeditor{jiachen}{JT}{0.0, 0.0, 0.9}
\addeditor{zhongpai}{ZP}{0.7, 0.0, 0.7}
\addeditor{ziyan}{ZW}{0.0, 0.5, 0.0}
\addeditor{benjamin}{BP}{0.8, 0.4, 0.1}
\addeditor{meng}{MZ}{0.5, 0.4, 0.1}
\addeditor{anwesa}{AC}{0.3, 0.8, 0.1}
\addeditor{nguyen}{VN}{0.9, 0.1, 0.9}
\showeditsfalse

% ---------------------------------------------------------------
% Title (using original arXiv title)
\title{From Particles to Fields:\\ Reframing Photon Mapping with Continuous Gaussian Photon Fields}

\titlerunning{Gaussian Photon Fields}

% Author list (from arXiv version, adapted for LNCS format)
\author{
Jiachen Tao\textsuperscript{$\dagger$}\inst{1,2} \and
Benjamin Planche\textsuperscript{$\ddagger$}\inst{2} \and
Van Nguyen Nguyen\textsuperscript{$\ddagger$}\inst{2} \and
Junyi Wu\textsuperscript{$\dagger$}\inst{1,2} \and
Yuchun Liu\inst{2} \and
Haoxuan Wang\inst{1} \and
Zhongpai Gao\inst{2} \and
Gengyu Zhang\inst{1} \and
Meng Zheng\inst{2} \and
Feiran Wang\inst{1} \and
Anwesa Choudhuri\inst{2} \and
Zhenghao Zhao\inst{1} \and
Weitai Kang\inst{1} \and
Terrence Chen\inst{2} \and
Yan Yan\inst{1} \and
Ziyan Wu\inst{2}
}

\authorrunning{J.~Tao et al.}

\institute{University of Illinois Chicago, Chicago, IL, USA \and
United Imaging Intelligence, Boston, MA, USA}

\maketitle

\begingroup
\renewcommand\thefootnote{$\dagger$}\footnotetext{This work was carried out during the internship of Jiachen Tao and Junyi Wu at United Imaging Intelligence, Boston MA, USA.}
\renewcommand\thefootnote{$\ddagger$}\footnotetext{Corresponding authors.}
\endgroup

\begin{abstract}
Accurately modeling light transport is essential for realistic image synthesis. 
Photon mapping provides physically grounded estimates of complex global illumination effects such as caustics and specular–diffuse interactions, yet its per-view radiance estimation remains computationally inefficient when rendering multiple views of the same scene. 
The inefficiency arises from independent photon tracing and stochastic kernel estimation at each viewpoint, leading to inevitable redundant computation.
To accelerate multi-view rendering, we reformulate photon mapping as a continuous and reusable radiance function. Specifically, 
we introduce the \textbf{Gaussian Photon Field (GPF)}, a learnable representation that encodes radiance distributions as anisotropic 3D Gaussian primitives parameterized by position, rotation, scale, and spectrum. 
GPF is initialized from physically-traced photons in the first SPPM iteration and optimized using multi-view supervision of final radiance, encoding photon-based light transport into a continuous field. 
After optimization, the field enables differentiable radiance evaluation along camera rays without repeated photon tracing or iterative refinement. 
Extensive experiments on scenes with complex light transport, such as caustics and specular–diffuse interactions, demonstrate that GPF attains photon-level accuracy while reducing computation by orders of magnitude, unifying the physical rigor of photon-based rendering with the efficiency of neural scene representations.
% TODO: Update keywords as needed
\end{abstract}

% ---------------------------------------------------------------
% Teaser figure (adapted for single-column ECCV format)
\begin{figure}[t]
    \centering
    \includegraphics[width=\linewidth, trim={4pt 4pt 4pt 4pt}, clip]{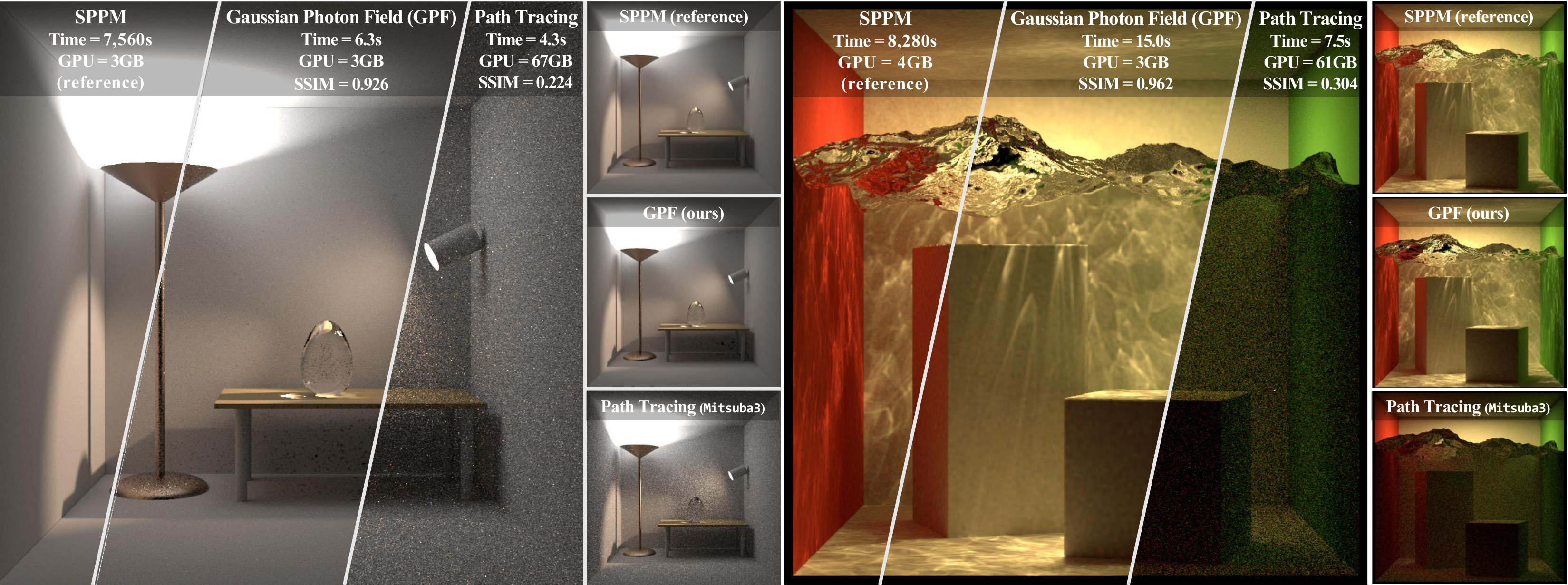}
    \caption{\textbf{Gaussian Photon Field (GPF)} unifies photon mapping and neural field representations,
        achieving physically accurate, efficient, and view-reusable global illumination across diverse scenes,
        including indirect lighting (\textit{left}) and water caustics (\textit{right}).
Using \textit{Stochastic Progressive Photon Mapping (SPPM)}~\cite{hachisuka2009sppm} and path tracing \cite{Mitsuba3} as reference, our method produces substantially higher structural similarity (SSIM), while reducing rendering costs by orders of magnitude compared to SPPM.
    }
    \label{fig:teaser}
    \vspace{-15pt}
\end{figure}

% ---------------------------------------------------------------
% Main content
\section{Introduction}
\label{sec:intro}
% 第一段不用变
Accurately modeling light transport \cite{kajiya1986rendering} is fundamental to realistic image synthesis, 
% which enables many applications such as AR/VR.
with applications in interactive graphics \cite{keller1997instant,zhang2019illumination,bitterli2020spatiotemporal}, medical imaging simulation \cite{wang1995mcml,fang2009monte,sarrut2022opengate}, scene understanding via inverse graphics \cite{barron2014shape,azinovic2019inverse,zhang2022modeling}, \etc.
Physically-based rendering techniques \cite{goral1984radiosity, veach1998bdpt}, such as photon mapping \cite{jensen1996photon} and its variants \cite{hachisuka2008progressive,hachisuka2009sppm,knaus2011probabilistic}, have long served as reliable estimators for complex global illumination, effectively capturing challenging phenomena such as caustics, indirect reflections, and refractions.
Despite their accuracy, these methods remain computationally expensive when rendering multiple views of the same scene. 

% 第二段应该只讲photon mapping是怎么解决complex global illumination的
Prior work on photon mapping \cite{jensen1996photon} addresses complex global illumination by decoupling light transport into two stochastic processes: photon tracing and density estimation.
During photon tracing, photons emitted from light sources are stored upon surface and volumetric interactions, forming a global photon map.
In the subsequent density estimation stage, the radiance at visible points is reconstructed by gathering nearby photons using spatial averaging based on stochastic kernel density estimation (KDE) \cite{silverman2018density,jensen2001realistic}.
This formulation enables accurate estimation of caustics and diffuse inter-reflections, making photon mapping a robust estimator for a wide range of lighting phenomena.

% 第三段应该去讲photon mapping在渲染一个场景的多视角的时候的问题

However, photon mapping is fundamentally ill-suited for multi-view rendering. While photon tracing is view-agnostic and the photon map can be reused, storing millions or billions of photons is often impractical. More importantly, the radiance-gathering step is inherently view-dependent \cite{jensen1996photon,hachisuka2009sppm,hachisuka2008progressive}, requiring recomputation for each camera view. The kernel density estimate introduces view-dependent bias, as each view gathers a different subset of photons, causing flickering, brightness shifts, or blur. Increasing photon density reduces variance but raises memory and computation costs. Thus, the tight coupling between visibility and per-view estimation makes classical photon mapping accurate but inefficient for multi-view reuse.

% 第四段应该去引入现代的neural radiance field是怎么用一个field去描述整个场景的信息并且可以在多视角渲染的时候reuse这个field。在photon mapping的过程中是xxxxx，收到neural radiance field的启发，我们是否可以用类似的field去学习整个场景的photon/radiance分布，这样子就可以在渲染新视角的时候xxxx
In contrast, modern neural scene representations, \eg, neural radiance fields (NeRF) \cite{nerf} and 3D Gaussian splatting (3DGS) \cite{3dgs}, represent scene appearance and geometry as continuous functions that can be queried from arbitrary viewpoints.
By learning such a shared representation, these methods amortize rendering costs across views.
%Inspired by this formulation, \textbf{we ask whether photon mapping—traditionally a discrete and view-dependent estimator—can be reformulated into a continuous radiance field that captures photon distributions throughout the scene}.
This motivates a fundamental question: \textbf{can photon mapping, traditionally a discrete, view-dependent estimator be reformulated as a continuous radiance field that encodes radiance distributions throughout the scene?}
Such a formulation would enable consistent photon reuse across views, preserving the physical accuracy of photon mapping while achieving neural-style efficiency.

% 第五段引入我们的方法，in this paper，we present gaussian photon field(GPF)，重点讲表征的设计和怎么用这套表征去渲染
%Building on this idea,
 Toward this goal, we introduce the \textbf{Gaussian Photon Field (GPF)}, a learnable radiance representation that unifies photon mapping and neural field paradigms.
% GPF models the global photon distribution as a set of anisotropic 3D Gaussian primitives, each encoding position, orientation, and directional radiance.
GPF models the global photon distribution as a set of anisotropic 3D Gaussian primitives, each parameterized by its position, rotation, scale, and spectrum, forming a continuous and differentiable radiance field that can be queried from arbitrary viewpoints. Unlike traditional 3D Gaussian splatting, which represents view-dependent surface appearance, GPF encodes view-independent radiance distributions that capture global light transport phenomena such as caustics and indirect illumination.
%\benjaminrmk{need to highlight fundamental differences with 3DGS}
% Once optimized, the field can be efficiently queried along camera rays to render novel views without re-tracing photons or performing kernel density estimation.
% Once optimized, the field can be efficiently queried along camera rays to render novel views \benjaminrmk{this repeats a bit the end of the previous sentence -- refactor a bit?}, eliminating the need for repeated photon tracing, stochastic density estimation, and progressive multi-pass refinement required by traditional photon mapping.
Once optimized, GPF enables efficient radiance evaluation along camera rays to render novel views, eliminating the need for repeated photon tracing, stochastic density estimation, or progressive refinement as in standard photon mapping.
Conceptually, GPF replaces per-view KDE with a single, continuous photon-based radiance field.

% 第六段讲我们怎么初始化和优化
We construct and optimize our Gaussian Photon Field in three stages:
(1) Initialization: Gaussian primitives are initialized from photons traced in a single iteration of stochastic progressive photon mapping (SPPM)~\cite{hachisuka2009sppm}, providing a physically grounded starting point;
(2) Radiance Query: a differentiable mechanism aggregates contributions from nearby Gaussians to estimate surface radiance, triggered at the first diffuse intersection after any specular or glossy bounces, at which point the ray terminates;
(3) Supervision: the field is optimized end-to-end against reference radiance computed offline by SPPM, using sparse multi-view supervision.
This process encodes complex, view-dependent light transport into a continuous, reusable radiance field.

% 第7段说to validate the effectiveness of our method xxxxxxx
To validate the effectiveness of our approach, we conduct extensive 
experiments across multiple physically-challenging scenes, including 
complex caustic and specular-diffuse interactions. We compare GPF against classical physically-based integrators (\eg, path 
tracing~\cite{Mitsuba3} and SPPM~\cite{hachisuka2009sppm}), Gaussian 
mixture fitting (EM-GMM~\cite{jakob2011progressive}), neural scene 
representations (\eg, NeRF~\cite{nerf}, 3DGS~\cite{3dgs}, and 
AnySplat~\cite{jiang2025anysplat}), and feed-forward rendering 
(RenderFormer~\cite{zeng2025renderformer}).
% Quantitative results are reported in terms of PSNR \cite{psnr}, SSIM \cite{ssim}, LPIPS (VGG) \cite{lpips}, along with render time and storage cost, while qualitative visualizations demonstrate that GPF achieves photon-level accuracy with significantly reduced computational redundancy. Together, these experiments 
Measuring visual quality, render time, and storage cost, results
confirm that GPF achieves photon-level accuracy with significantly reduced computational redundancy and successfully unifies the physical accuracy of photon-based rendering with the efficiency and scalability of continuous learned representations.

\section{Related Work}
\label{sec:related_work}

% Our work, Gaussian Photon Fields (GPF), creates a bridge between classical physically-based rendering and modern neural representations. Its novelty is best understood by contextualizing it against these two distinct fields, as well as prior work on radiance caching.

\noindent\textbf{Physically-Based Global Illumination.}
\label{sec:related_physically_based}
Simulating global illumination (GI) \cite{kajiya1986rendering} is a central problem in computer graphics. Unbiased Monte Carlo methods, \eg, path tracing (PT)~\cite{kajiya1986rendering} and its bidirectional variants (BDPT)~\cite{veach1998bdpt}, are considered the gold standard for accuracy. However, they struggle to efficiently sample complex, low-probability light paths, such as those that form caustics, often resulting in high variance and noise.

\begin{wrapfigure}{r}{0.5\textwidth}
    \vspace{-15pt}
    \centering
    \includegraphics[width=0.48\textwidth]{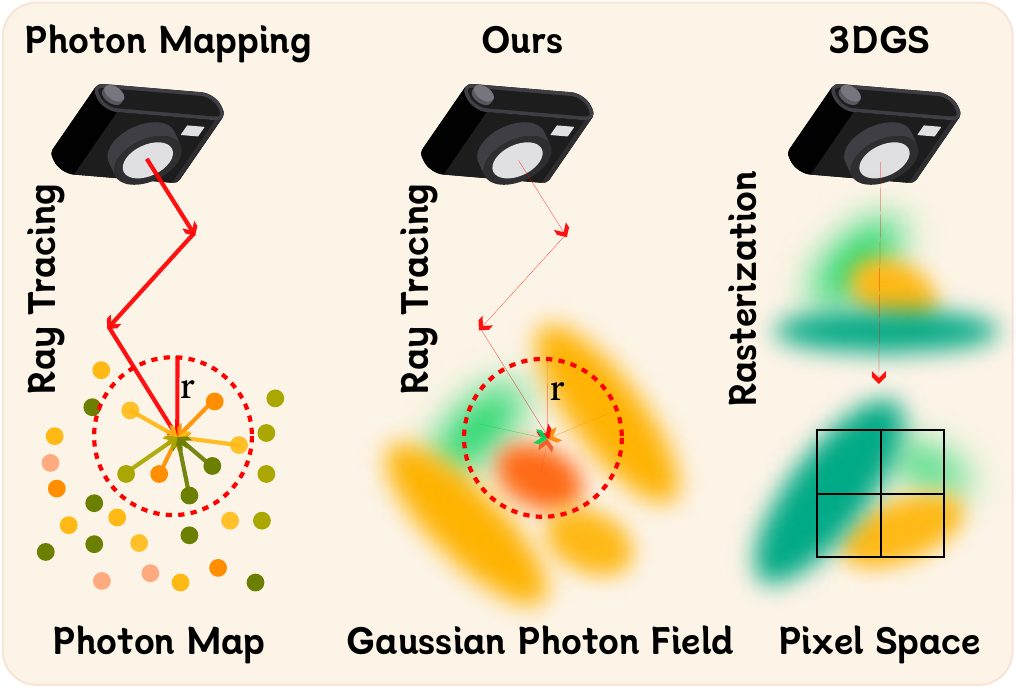}
    \caption{\textbf{Rendering paradigm comparison.}
    Classical photon mapping estimates radiance by gathering discrete photons within a local radius~$r$ during ray tracing.  GPF replaces this with a continuous, differentiable photon field queried via ray tracing. In contrast, 3DGS rasterizes  primitives in pixel space, emphasizing appearance over physically-grounded light transport.}
    \label{fig:method_compare}
    \vspace{-15pt}
\end{wrapfigure}

To address this, biased techniques like photon mapping~\cite{jensen1996photon} and its stochastic progressive variant SPPM~\cite{hachisuka2009sppm} were developed. As detailed in Sec.~\ref{sec:preliminaries}, these methods excel at capturing complex GI phenomena by decoupling light transport into a two-pass process. Their primary limitation, however, is the \textit{per-view estimation paradigm}, which fundamentally relies on a view-dependent kernel density estimation (KDE) process that must be re-executed for every new camera view. Our work directly targets this bottleneck. As illustrated in Figure~\ref{fig:method_compare}, instead of proposing a new sampling strategy for these methods, we reformulate their core estimation process, replacing the repeated, per-view KDE with a query to a persistent, continuous radiance function.

\noindent\textbf{Neural and Learned Scene Representations.}
\label{sec:related_neural}
The concept of a continuous, learnable scene representation was popularized by neural radiance fields (NeRF)~\cite{nerf}. NeRF and its many successors~\cite{chen2024narcan,guo2023forwardflow,li2021nsff,liu2023robustdynrf,ma2024humannerfse,park2021nerfies,park2021hypernerf,pumarola2020dnerf,tretschk2021nonrigid,wu2025denver,xian2021stnerf} achieve state-of-the-art novel view synthesis. However, these methods are essentially ``black-box" appearance models: optimized for view interpolation, they lack physical grounding and cannot guarantee accurate light transport for complex effects such as caustics or relighting.

More recently, 3D Gaussian splatting (3DGS)~\cite{3dgs,2dgs,yang2024specgaussian} has achieved remarkable success, combining the efficiency of explicit primitives with the representational quality of neural scene models for real-time reconstruction. We draw inspiration from 3DGS in its use of Gaussians as compact, expressive primitives. However, our approach differs fundamentally in both objective and rendering process. While 3DGS models scene appearance and geometry through differentiable rasterization (splatting), our photon field represents the scene’s global illumination and radiance distribution arising from physical light transport. As shown in Figure~\ref{fig:method_compare}, instead of rasterizing Gaussians to reproduce appearance, we employ them within a differentiable ray-tracing framework, where radiance is computed through continuous field queries. GPF thus extends Gaussian primitives to a physically-grounded domain, enabling efficient and reusable global illumination estimation across views.

\noindent\textbf{Density Learning with Gaussian Mixtures.
}
\label{sec:related_caching}
% Radiance caching techniques~\cite{placeholder_ward88_caching, caching2} 
% precompute indirect illumination to accelerate rendering. Neural 
% Radiance Caching~\cite{M_ller_2021} extends this for real-time 
% performance, but relies on path tracing which struggles with caustics. 
% GPF instead leverages photon mapping for accurate caustics, building 
% a persistent representation that amortizes light transport across views.
Gaussian mixture models (GMM)~\cite{condor2025dont,Xu13sigasia,green2006view,vorba2014line,yan2016position,jakob2011progressive,zhou2024unifiedgaussianprimitivesscene} have also been used to approximate light transport. 
Jakob \etal~\cite{jakob2011progressive} proposed using Expectation-Maximization 
(EM) to fit Gaussian mixtures to photon distributions in \textit{volumetric} 
participating media. Their derivation relies on a key property: a 3D Gaussian 
integrates to unity over 3D space, \ie, $\int G(\mathbf{x}) dV = 1$, enabling 
closed-form density estimation.
However, this formulation does not transfer to \textit{surface} photon mapping. 
For photons deposited on 2D manifolds, kernel density estimation requires 
$\int K(\mathbf{x}) dS = 1$ over the surface. A 3D Gaussian does not satisfy 
this constraint, its integral over an arbitrary surface $S$ depends on the 
position, orientation, and local geometry of $S$ relative to the Gaussian, 
with no closed-form solution. Fitting 2D Gaussians in a global UV parameterization 
is also impractical for complex geometry with discontinuous or overlapping 
parameterizations.
Our method differs fundamentally in 2 aspects:
(1)~\textbf{Learning target}: EM-GMM fits photon \textit{density} and analytically 
derives radiance; GPF directly learns \textit{radiance} via supervision.
(2)~\textbf{Optimization}: EM-GMM uses unsupervised EM; GPF uses gradient-based 
optimization with multi-view radiance supervision.
% (3)~\textbf{Representation}: While we use Gaussians for efficiency, this is a 
% design choice—not a mathematical necessity as in GMM. A NeRF-based representation could equivalently
% serve as the radiance representation.
\section{Preliminaries}
\label{sec:preliminaries}

Our work bridges classical light transport simulation with modern, learnable representations. To formally ground our method, we first review the rendering equation and then summarize the mathematics of progressive photon mapping---the paradigm we aim to reformulate.

\subsection{Rendering Equation}
Physically-based rendering aims to solve the rendering equation~\cite{kajiya1986rendering}, a formulation of light transport in equilibrium. It states that the outgoing radiance $L_o$ 
% (light leaving) 
%
% \begin{equation}
% L_o(\mathbf{x}, \omega_o)
% = L_e(\mathbf{x}, \omega_o)
% + \int_{\Omega} f_r(\mathbf{x}, \omega_i, \omega_o)
%   L_i(\mathbf{x}, \omega_i)\,
%   (\mathbf{n}\cdot\omega_i)\, d\omega_i.
% \end{equation}
from a surface point $\mathbf{x}$ in a direction $\omega_o$ is the sum of its emitted light $L_e$ and all reflected incoming light:
\begin{equation}
L_o(\mathbf{x}, \omega_o) 
= L_e(\mathbf{x}, \omega_o) 
+ \int_{\Omega} f_r\big(\mathbf{x}, (\substack{\omega_i\\\omega_o})\big)\, 
  L_i(\mathbf{x}, \omega_i)\, (\mathbf{n}\cdot\omega_i)\, d\omega_i,
  \label{eq:rendering_equation}
\end{equation}
where $f_r$ is the bidirectional reflectance distribution function (BRDF) \cite{nicodemus1965directional} describing the material, $\mathbf{n}$ is the surface normal, $\Omega$ is the hemisphere of all incoming directions $\omega_i$, and $L_i$ is the incoming radiance 
%(light arriving) 
at $\mathbf{x}$ from direction $\omega_i$.

The primary challenge in rendering is solving for $L_i$, as it is the solution to another rendering equation at the next surface intersection. This recursive integral is particularly difficult to solve for complex light paths, such as caustics, where high-energy contributions from a light source are focused onto a diffuse surface after interacting with a specular (\eg, glass or metal) surface. Unidirectional Monte-Carlo methods, \eg, path tracing~\cite{kajiya1986rendering}, struggle to sample these low-probability paths efficiently, leading to high variance.

\subsection{Stochastic Progressive Photon Mapping}
Photon mapping~\cite{jensen1996photon} and its progressive variants~\cite{hachisuka2008progressive,hachisuka2009sppm} were introduced specifically to solve these difficult light paths. SPPM reformulates the problem by decoupling light transport into two passes.

\noindent\textbf{Pass 1: Photon Tracing.} In a view-independent pass, photons are traced from light sources till their interaction with non-specular surfaces. Each photon $p$ records its position $\mathbf{x}_p$ and flux $\Delta\Phi_p$, stored in a global spatial structure, \eg, a k-d tree or hash grid, forming the photon map for subsequent density estimation.

\noindent\textbf{Pass 2: Radiance Estimation.} In a view-dependent pass, radiance is computed at all visible surface points $\{\mathbf{x}\}$ for the current camera. SPPM performs this using iterative kernel density estimation. In each iteration $k$, a search radius $r_k$ is used to find nearby photons. Radiance is estimated by summing the flux of all $N_p$ photons within the radius and normalizing by the kernel area:
\begin{equation}
    L_i(\mathbf{x}) \approx \frac{1}{\pi r_k^2} \sum_{p=1}^{N_p} \Delta\Phi_p.
    \label{eq:kde}
\end{equation}
This radiance estimate is progressively refined by tracing more photons (increasing $N_p$) and shrinking the radius $r_k$, which allows the estimate to converge to a physically accurate result.
Crucially, 
% as we established in Section~1,
% as established in the Introduction,
this process of gathering photons at a visible point and managing its local radius is fundamentally coupled to the camera view. 
% This section provides the mathematical basis for this \textbf{per-view estimation paradigm}, which our method will replace with a continuous, reusable function.
It underpins the \textbf{per-view estimation paradigm} that limits photon mapping, whereas our method replaces it with a continuous, reusable function.
\begin{figure*}[!t]
    \centering
    \includegraphics[width=\textwidth]{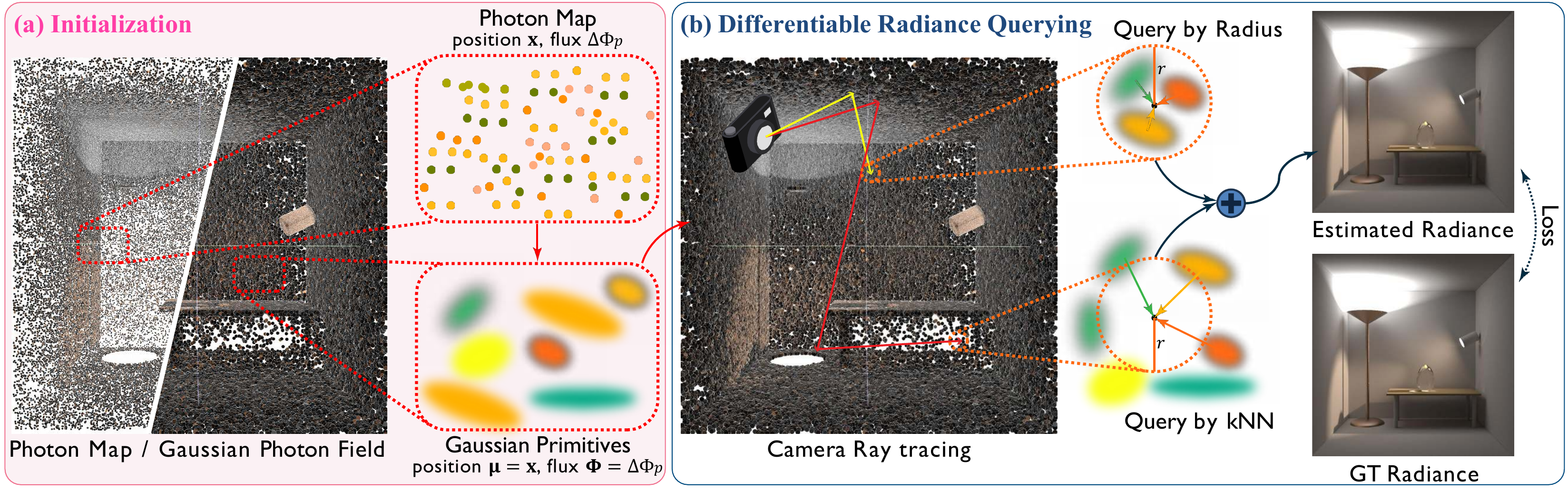}
    \caption{
    \textbf{Overview of the Gaussian Photon Field (GPF) pipeline.}
    ~\textbf{\textcolor{MyPink}{(a) Initialization.}}
    We convert the discrete photon map from one SPPM iteration into a continuous Gaussian photon field by instantiating one anisotropic 3D Gaussian per photon. Each primitive inherits the photon's position and flux, with rotation and scale randomly initialized.
    ~\textbf{\textcolor{figure3b}{(b) Differentiable Radiance Querying.}}
    During training and rendering, camera rays are traced through the scene. At the first diffuse intersection, radiance is estimated by aggregating contributions from nearby Gaussians via a radius-based and k-nearest-neighbor query. The predicted radiance is supervised by the GT values at corresponding surface points, enabling end-to-end optimization of all Gaussian parameters and producing a differentiable, view-reusable radiance field.
    }
    \label{fig:pipeline}
    \vspace{-16pt}
\end{figure*}

\section{Method}
\label{sec:method}

\subsection{Gaussian Photon Field Representation} \label{sec:method_representation}

Our method begins by recasting the discrete particle set of SPPM into a continuous, differentiable representation. Instead of a sparse point cloud, we introduce a Gaussian photon field (GPF), defined as a mixture of $N$ anisotropic 3D Gaussian primitives,
$\{\mathcal{G}_i\}_{i=1}^N$.
Each primitive is parameterized as $\mathcal{G}_i = (\boldsymbol{\mu}_i,\mathbf{q}_i,\mathbf{s}_i,\boldsymbol{\Phi}_i)$, with a mean position $\boldsymbol{\mu}_i \in \mathbb{R}^3$, a rotation $\mathbf{q}_i \in \mathbb{H}$ (represented as a unit quaternion), a scale $\mathbf{s}_i \in \mathbb{R}^3_+$, and a spectral flux $\boldsymbol{\Phi}_i \in \mathbb{R}^3$. The rotation and scale collectively define the anisotropic covariance matrix $\boldsymbol{\Sigma}_i$. The radiance contribution $G_i$ of a single primitive at any 3D point $\mathbf{x}$ is then given by its un-normalized, flux-weighted Gaussian function: 
\begin{equation} G_i(\mathbf{x}) = \boldsymbol{\Phi}_i \exp\!\Big[-\tfrac{1}{2}(\mathbf{x}-\boldsymbol{\mu}_i)^\top \boldsymbol{\Sigma}_i^{-1}(\mathbf{x}-\boldsymbol{\mu}_i)\Big]. \label{eq:gaussian_contribution} 
\end{equation} 
A key component of our method is its \textbf{physically-inspired initialization}. The GPF is not created from scratch, but directly seeded from classical simulation. We run a single iteration of SPPM to trace and store a sparse photon map. For each of the $N$ photons $p$ in this map, we instantiate one Gaussian primitive $\mathcal{G}_i$, mapping the physical properties directly.
% \begin{itemize} \item \textbf{Position:} 
The Gaussian mean is set to the photon's surface position: $\boldsymbol{\mu}_i \leftarrow \mathbf{x}_p$,  
% \textbf{Spectral Flux:} 
and the Gaussian flux is set to the photon's flux: $\boldsymbol{\Phi}_i \leftarrow \Delta\Phi_p$. 
% \textbf{Shape:} 
Rotation $\mathbf{q}_i$ and scale $\mathbf{s}_i$ have no direct analog in a discrete particle. We therefore randomly initialize $\mathbf{q}_i$ and set the initial $\mathbf{s}_i$ to a small, isotropic value, providing diverse local support for subsequent optimization. Through gradient-based optimization, orientations and scales will align with dominant light paths.
% \end{itemize} 
% This "particle-to-field" process is illustrated in \autoref{fig:pipeline}: (a) sparse photons from the first SPPM iteration; initialization as isotropic Gaussian ellipsoids at the photon locations; (b) the final optimized Gaussians, whose orientations and scales have aligned with dominant light paths. This approach transforms the sparse, discrete photon set into a continuous, differentiable field, setting the stage for gradient-based optimization and efficient, reusable radiance queries.
This \textit{particle-to-field} process (illustrated in Figure~\ref{fig:pipeline}) results in a continuous, differentiable field, setting the stage for gradient-based optimization and efficient, reusable radiance queries.

\subsection{Differentiable Radiance Querying}
\label{sec:camera_ray_tracing}

During training and inference, rendering proceeds by tracing camera rays through the scene for radiance estimation, querying the Gaussian photon field, which encodes the spatial distribution of photon energy. 
% At each visible surface point, the incoming radiance is obtained by querying the continuous Gaussian Photon Field, 
% which encodes the spatial distribution of photon energy learned during training. 
This design replaces discrete photon-based estimation with a continuous radiance function, 
allowing radiance to be evaluated directly from the learned Gaussian field without repeated tracing.
This section describes our hybrid ray tracing pipeline (Sec.~\ref{sec:ray_tracing}) and the radiance query mechanism (Sec.~\ref{sec:radiance_query}).

\noindent\textbf{
Ray Tracing with Gaussian Radiance Integration.
}
\label{sec:ray_tracing}
We employ a path-tracing–style camera integrator augmented with Gaussian radiance field queries for diffuse interactions. 
Given a camera ray $\mathbf{r}(t) = \mathbf{o} + t\mathbf{d}$, the outgoing radiance is computed by the rendering equation:
% \begin{equation}
%     L_o(\mathbf{x}, \omega_o) = L_e(\mathbf{x}, \omega_o) + 
%     \int_{\Omega} f_s(\mathbf{x}, \omega_i, \omega_o) 
%     L_i(\mathbf{x}, \omega_i) |\cos\theta_i| \, d\omega_i,
% \end{equation}
\begin{equation}
    L_o(\mathbf{x}, \omega_o) = L_e(\mathbf{x}, \omega_o) \\
    + 
    \int_{\Omega} f_r\big(\mathbf{x}, (\substack{\omega_i\\\omega_o})\big) 
    L_i(\mathbf{x}, \omega_i) |\cos\theta_i| \, d\omega_i,
\end{equation}
where $L_i(\mathbf{x}, \omega_i)$ is approximated by 
$L_{\text{GPF}}(\mathbf{x}, \omega_i)$.

Our algorithm follows the standard path-tracing loop but introduces a key modification at diffuse interactions.
When a view ray first encounters a diffuse surface, 
instead of spawning further secondary rays or gathering photons, 
we query GPF to obtain the preaccumulated incident radiance.
This hybrid design drastically reduces variance and avoids double-counting BSDF contributions, 
as the radiance stored in the field already encodes the expected reflected contribution from photon sampling during optimization. 
The accumulated radiance at a diffuse surface point $\mathbf{x}$ is thus:
\begin{equation}
    L(\mathbf{x}) = L_e(\mathbf{x}) + \beta \, L_\text{GPF}(\mathbf{x}),
\end{equation}
where $\beta$ denotes the current throughput. 
For specular or glossy interactions with probability $p(\omega_o)$ of sampling the direction $\omega_o$, we continue the path tracing as:
\begin{equation}
    \beta \leftarrow \beta \cdot 
    \frac{f_r\big(\mathbf{x}, (\substack{\omega_i\\\omega_o})\big) |\cos\theta_i|}{p(\omega_o)}.
\end{equation}
This hybrid formulation preserves physical correctness for specular transport while amortizing diffuse transport via the learned Gaussian radiance. As shown in Figure~\ref{fig:pipeline}, camera rays propagate through the scene: specular and glossy bounces use classical path tracing, while diffuse bounces query the Gaussian field for fast radiance estimation.

\noindent\textbf{Radiance Query and Aggregation.}
\label{sec:radiance_query}
% The Gaussian photon field represents global illumination as a set of $N$ anisotropic Gaussian primitives
% \[
% \mathcal{G}=\{(\boldsymbol{\mu}_i,\mathbf{q}_i,\mathbf{s}_i,\boldsymbol{\Phi}_i)\}_{i=1}^{N},
% \]
% where $\boldsymbol{\mu}_i\in\mathbb{R}^3$ is the center, $\mathbf{q}_i\in\mathbb{H}$ is a unit quaternion parameterizing rotation, 
% $\mathbf{s}_i\in\mathbb{R}^3_{+}$ are per-axis scales, and $\boldsymbol{\Phi}_i\in\mathbb{R}^3$ is the learned spectrum (radiance/flux).
% Let $\mathbf{R}_i=\mathbf{R}(\mathbf{q}_i)\in SO(3)$ denote the rotation matrix induced by $\mathbf{q}_i$, and 
% $\mathbf{D}_i=\mathrm{diag}(\mathbf{s}_i)$ the scale matrix. The (implicit) precision of the $i$-th Gaussian is then
% $\mathbf{\Lambda}_i=\mathbf{R}_i\,\mathbf{D}_i^{-2}\,\mathbf{R}_i^{\top}$. \benjaminrmk{why is the Gaussian definition repeated here? Can we delete the whole paragraph?}
Given a surface query point $\mathbf{x}$ (first diffuse hit) with outgoing direction $\omega_o$ toward the viewer, we gather contributions from a spatial neighborhood $\mathcal{N}(\mathbf{x})$ of nearby Gaussians.
% (radius query with $r$; if $|\mathcal{N}|<k_{\min}$, we augment with $k$-NN \benjaminrmk{we should probably develop this and explain why, \cf richer backpropagation?}).
The GPF aggregated radiance is computed as:
% \begin{equation}
\begin{align}
L_{\mathrm{GPF}}(\mathbf{x}, \omega_o)
&= \sum_{i\in\mathcal{N}(\mathbf{x})}
w_i(\mathbf{x})\,\boldsymbol{\Phi}_i,
% \end{equation}
% where $f_r(\mathbf{x}, \omega_o)$ is the surface BSDF evaluated at the query point, enabling view-dependent shading from view-independent Gaussian primitives. \benjaminrmk{shouldn't $f_r$ take 2 directions as parameters, as in Equation \ref{eq:rendering_equation}? If so, it should be corrected in the 2 paragraphs below too.}
% The spatial weight $w_i(\mathbf{x})$ is an anisotropic Gaussian kernel modulated by a smooth distance falloff:
% \begin{equation}
\\
    \text{with } w_i(\mathbf{x})
    &= \exp{\!\Big(-\tfrac{1}{2}\,(\mathbf{x}-\boldsymbol{\mu}_i)^{\top}
                  \mathbf{\Lambda}_i\,(\mathbf{x}-\boldsymbol{\mu}_i)\Big)}
      \;\cdot\;
      \psi\!\big(\|\mathbf{x}-\boldsymbol{\mu}_i\|\big),
% \end{equation}
\end{align}
\ie, with $w_i(\mathbf{x})$ an anisotropic Gaussian kernel modulated by a smooth distance falloff, as spatial weight ; and $\psi(\cdot)$ a soft attenuation that decays near a maximum radius $r_{\max}$ to ensure stability and avoid hard boundaries. The BSDF term 
% $f_r(\mathbf{x}, \omega_o)$ 
$f_r$
modulates the field query by the surface reflectance properties, ensuring that Gaussian primitives $\boldsymbol{\Phi}_i$ encode a base radiance field that is view-independent, while the final radiance $L_{\mathrm{GPF}}$ correctly accounts for directional reflectance. All terms are differentiable \wrt\ $(\boldsymbol{\mu}_i,\mathbf{q}_i,\mathbf{s}_i,\boldsymbol{\Phi}_i)$, enabling end-to-end optimization.

\noindent\textbf{Neighborhood Search.}
To balance efficiency and adaptivity, we employ a hybrid spatial retrieval strategy: % to define $\mathcal{N}(\mathbf{x})$:
\begin{equation}
    \mathcal{N}(\mathbf{x}) =
    \text{BallQuery}(\mathbf{x}, r) \cup 
    \text{kNN}\big(\mathbf{x}, \max(0, k_\text{min} - k_r)\big).
\end{equation}
\Ie, for dense photon regions, we perform a radius-based query using a KD-tree ($\text{BallQuery}$), returning $k_r$ Gaussians. 
In sparse regions, where $k_r < k_\text{min}$ neighbors are found, 
we supplement with a k-nearest-neighbor ($\text{kNN}$) query to guarantee coverage. 
This ensures consistent density support while maintaining smooth transitions.

\noindent\textbf{Differentiable Accumulation.}
All query components---including anisotropic weighting, soft distance attenuation, and normalization---are differentiable \wrt Gaussian parameters
$(\boldsymbol{\mu},\mathbf{q},\mathbf{s},\boldsymbol{\Phi})$.
Implementation relies on \texttt{Dr.Jit}~\cite{Jakob2022DrJit} with full GPU vectorization, aggregating per-neighborhood contributions via a parallel scatter-reduce.
This enables efficient batched training and stable gradient-based optimization, supporting fast radiance evaluation once the field is trained without relying on non-differentiable lookups or per-view kernel density estimation.

\subsection{Supervision and Training}
\label{sec:supervision_training}

The Gaussian photon field is trained under sparse multi-view supervision derived from physically-based progressive photon mapping~\cite{hachisuka2009sppm}. 
Instead of supervising at the image level, we operate at the level of \textit{first-bounce diffuse surface points}---that is, the surface intersections where camera rays first hit diffuse materials. 
This design ensures precise geometric alignment between the queried radiance and ground-truth locations, while excluding background and invalid samples.

\noindent\textbf{Ground-truth Generation.}
For each of the $k$ training camera views, we gather the set of diffuse surface points visible from the camera:
% \begin{equation}
% \begin{split}
% \mathcal V
% = \bigl\{&\mathbf{x}_i \in \mathcal D \,\big|\,
% \exists \gamma \in \mathcal P(o,\mathbf{x}_i) : \\
% &\text{NonDiffuse}(\gamma,\mathcal S\setminus \mathcal D) \;\wedge\;
% \text{FirstDiffuse}(\gamma,\mathcal D) \,\bigr\},
% \end{split}
% \label{eq:visible_diffuse}
% \end{equation}
$
\mathcal V
= \bigl\{\mathbf{x}_i \in \mathcal D \,\big|\,
\exists \gamma \in \mathcal P(o,\mathbf{x}_i) : 
\text{NonDiffuse}(\gamma,\mathcal S\setminus \mathcal D) \;\wedge\;
\text{FirstDiffuse}(\gamma,\mathcal D) \,\bigr\},
$
where $o$ is the camera position, $\mathcal{S}$ is the full scene geometry, $\mathcal{D} \subset \mathcal{S}$ is the diffuse subset, and $\mathcal{P}(o,\mathbf{x})$ denotes the set of all finite piecewise-linear paths from $o$ to $\mathbf{x}$. The predicate $\text{NonDiffuse}(\gamma, \mathcal{S}\setminus\mathcal{D})$ is true if each intermediate point of $\gamma$ lies on a non-diffuse surface, including both specular and glossy interactions, and $\text{FirstDiffuse}(\gamma, \mathcal{D})$ is true if the path first intersects a diffuse point at its endpoint.
% } \benjaminrmk{not sure if better than the non-formal version below, but sharing just in case...}
% For each of the $k$ training camera views, we trace primary rays and record their first intersections with diffuse surfaces:
% \begin{equation}
% \mathcal{V} = \{\, \mathbf{x}_i \mid \text{camera ray first hits a diffuse surface point} \,\}.
% \end{equation}

% For every supervision point $\mathbf{x}\in\mathcal{V}$, the ground-truth radiance is computed using a high-quality Stochastic Progressive Photon Mapping (SPPM) reference.
For every supervision point $\mathbf{x}\in\mathcal{V}$, the ground-truth radiance is computed using a high-quality Stochastic Progressive Photon Mapping (SPPM) reference. Although generating these references is computationally expensive, it is performed only once for training; once optimized, our field enables fast, reusable radiance queries for arbitrary novel views.
% \benjaminrmk{I feel readers might wonder about the costs of generating the GT images ; we might want to further highlight the benefits of our method for novel view synthesis (\ie, our method is \textit{maybe a bit costly} to train but benefits downstream applications)...?}
Following the standard SPPM procedure, photons are traced from light sources and stored only on diffuse surfaces during each iteration.
The radiance at $\mathbf{x}$ is then estimated by gathering nearby photons and averaging across $N_{\text{iter}}$ progressive iterations:
\begin{equation}\label{eq:gt_radiance}
L_{\text{ref}}(\mathbf{x}, \omega_o) = \frac{1}{N_{\text{iter}}} \sum_{t=1}^{N_{\text{iter}}} \frac{1}{\pi r_t^2} \sum_{\mathbf{p}_i \in \mathcal{B}_t(\mathbf{x})} \Phi_i \cdot f_r(\mathbf{x}, \omega_i, \omega_o),
\end{equation}
where $\omega_i$ is the photon incident direction, $\omega_o$ the camera viewing direction, $\mathcal{B}_t(\mathbf{x})$ the photon neighborhood within radius $r_t$, $\Phi_i$ the photon power, and $f_r$ the BSDF~\cite{nicodemus1965directional,hachisuka2009sppm}.
This yields a low-variance, physically-accurate estimate of the directional outgoing radiance at $\mathbf{x}$.
% All reference radiance values are stored in linear RGB space as raw \texttt{.npy} arrays and directly used as point-wise supervision for training the Gaussian Photon Field.
% \benjamin{All reference radiance values are pre-computed and stored in linear RGB space, before being used as point-wise supervision for training GPF.}
% \jiachen{Actually all reference radiance values are not in RGB space, I guess they are in HDR space(I need check)}

\noindent\textbf{Training.}
During optimization, we trace camera rays for the same $k$ reference views and query the photon field at each \textit{first diffuse surface intersection} to obtain predicted radiance $L_{\text{GPF}}(\mathbf{x})$. 
The field parameters 
% $\Theta = \{ \mathbf{c}_i, \mathbf{R}_i, \mathbf{s}_i, \boldsymbol{\phi}_i \}$ 
are optimized by minimizing a mean-squared error (MSE) loss in radiance space:
\begin{equation}
\mathcal{L}_{\text{MSE}} = \frac{1}{|\mathcal{V}|}
\sum_{\mathbf{x}\in\mathcal{V}}
\| L_{\mathrm{GPF}}(\mathbf{x}, \omega_o) - L_{\text{ref}}(\mathbf{x}, \omega_o) \|_2^2.
\end{equation}
% Gradients are back-propagated through both the Gaussian query and the camera ray-tracing process in a fully differentiable manner.

\section{Experiments}
\label{sec:experiments}
% In this section, we first describe the experimental setup (Sec.~\ref{sec:exp_setup}). We then compare quantitatively and qualitatively our method with both classical physically-based integrators (path tracing~\cite{Mitsuba3} and SPPM~\cite{hachisuka2009sppm}) and standard neural rendering approaches (Instant-NGP~\cite{mueller2022instant} and 3D Gaussian Splatting~\cite{3dgs}) using our newly proposed challenging datasets~(Sec.~\ref{sec:exp_quantitative} and Sec.~\ref{sec:exp_qualitative}). Finally, we present ablation studies to evaluate the impact of various design choices in our approach (Sec.~\ref{sec:exp_ablation}).
We validate our method on our newly-proposed challenging benchmark, comparing to both classical physically-based integrators and standard neural rendering approaches and justifying design choices via multiple ablation studies.

\subsection{Experimental Setup}
\label{sec:exp_setup}

% \noindent\textbf{Datasets.} We evaluate our method on five physically based rendering scenes that cover a wide
% range of light–transport phenomena, including specular–diffuse interactions,
% caustics, and glossy reflections:

% \begin{itemize}
%   \item \textbf{Veach-Bidir} and \textbf{Veach-Ajar}, both classic test scenes from Benedikt~Bitterli’s rendering resources~\cite{resources16},
%   featuring multi-bounce specular–diffuse paths and complex indirect illumination.
%   We use their original geometry and materials, converted to Mitsuba 3\cite{Mitsuba3} format.

%   \item \textbf{Water-Caustic} and \textbf{Water-Caustic~2}, scenes inspired by Bitterli’s dataset\cite{resources16} but reconfigured for Mitsuba 3\cite{Mitsuba3} due to the lack of suitable XML definitions.

%   \item \textbf{Cornell Box}\cite{Mitsuba3}, a standard diffuse reference scene used to evaluate
%   overall radiometric accuracy and convergence behavior.
\noindent\textbf{Datasets.} Since no standard or well-established datasets exist for this task, we propose a new one consisting of five physically-based rendering scenes that cover a wide range of light-transport phenomena, including specular–diffuse interactions, caustics, and glossy reflections:
\begin{itemize}
  \item \textit{Veach-Bidir} and \textit{Veach-Ajar}: classic Mitsuba-3-compatible test scenes from Bitterli’s rendering resources~\cite{resources16},
  featuring multi-bounce specular–diffuse paths and complex indirect illumination.
  % We use their original geometry and materials, converted to ~\cite{Mitsuba3} format.
  \item \textit{Water-Caustic} and \textit{Water-Caustic~2}: scenes inspired by \citet{resources16} and reconfigured for Mitsuba 3, with challenging refractive and specular-diffuse caustics.
  % Bitterli’s dataset 
  %due to the lack of suitable XML definitions.
  \item \textit{Cornell Box}~\cite{Mitsuba3}: canonical diffuse reference scene to evaluate
  radiometric accuracy and convergence behavior. 
\end{itemize}

% \noindent\textbf{Baselines.} \nguyen{ We compare our proposed method, GPF, with both unbiased and photon-based rendering techniques to evaluate its accuracy, efficiency, and multi-view consistency. Specifically, we benchmark GPF against standard Monte Carlo path tracing~\cite{} and Stochastic Progressive Photon Mapping (SPPM)~\cite{} under comparable conditions. For the unbiased baseline, we use the path tracing integrator from Mitsuba 3~\cite{Mitsuba3} at two sampling budgets (10 and 50~spp), representing classical Monte Carlo rendering at different noise levels. For the photon-based baseline, we use SPPM as both a baseline and a high-quality reference: \textit{SPPM - 3 iters} serves as a time-matched baseline to GPF with comparable render time, while \textit{SPPM - 1000 iters} provides the ground-truth reference for all quantitative and qualitative evaluations.}

\noindent\textbf{Training Protocol.}
% For the \textit{Veach–Bidir} scene, GPF is trained on 30 views using per-pixel final radiance from SPPM. For the other scenes, the model relies on only 3 views.
% Radiance maps are rendered at a resolution of $720\times1280$ for \textit{Veach–Ajar}, and $1024\times1024$ for the rest.
The model is trained with per-pixel final radiance from SPPM, using only 3 views and $1024\times1024$ radiance maps for all scenes except \textit{Veach–Bidir} (30 views, $720\times1280$ maps). 
%More details are provided in the supplementary material.
See the appendix for more details.

% \noindent\textbf{Baselines and reference.}
% We compare GPF against Mitsuba 3 \textbf{path tracing} integrator\cite{Mitsuba3} at two sampling
% budgets (10 and 50~spp), representing classical unbiased Monte Carlo rendering
% at different noise levels. In addition, we use Stochastic Progressive Photon
% Mapping (SPPM) both as a baseline and as a high-quality reference:
% \textbf{SPPM - 3 iters} serves as a time-matched baseline to GPF with comparable
% render time, while \textbf{SPPM - 1000 iters} provides the ground-truth reference for
% all quantitative and qualitative evaluations. This configuration allows us to
% fairly assess GPF’s rendering accuracy, efficiency, and multi-view consistency
% against both unbiased and photon-based estimators.

\subsection{Quantitative Comparisons and Ablation Studies}
\label{sec:exp_quantitative}

\noindent\textbf{Comparison with Baselines.} 
Table~\ref{tab:whole_scene_tab} compares GPF against physically-based 
integrators (path/particle tracing~\cite{Mitsuba3}, SPPM~\cite{hachisuka2009sppm}), 
Gaussian mixture modeling (EM-GMM \cite{jakob2011progressive}), and 
feed-forward neural rendering (RenderFormer~\cite{zeng2025renderformer}). 
For EM-GMM, we reproduce the method in Mitsuba 3 as no public code is 
available. 
For RenderFormer, we use the official code and convert our scenes to 
their format.
% Each scene targets a distinct class of light transport effects:
% \textit{Veach--Bidir} and \textit{Veach--Ajar} emphasize complex indirect illumination strong multi-bounce diffuse transport;
% \textit{Water--Caustic} and \textit{Water--Caustic~2} focus on challenging refractive and specular-diffuse caustics;
% and the \textit{Cornell Box} provides a canonical test for general global illumination.
% \benjaminrmk{deleted paragraph redundant with Datasets paragraph.}

\begin{table*}[t]
    \setlength{\tabcolsep}{2.5pt}
    \centering
    \small
    \caption{\textbf{Comparison across 5 complex scenes}, reporting PSNR, SSIM, LPIPS, and storage cost (in MB). \colorbox{red!25}{Best} and \colorbox{orange!25}{\nth{2} best} results are highlighted.
    % \benjaminrmk{suggestion: count the storage size of the meshes for all methods that rely on them during rendering.}
    }
    \vspace{-5px}
    \label{tab:whole_scene_tab}
    \resizebox{\textwidth}{!}{%
    \begin{tabular}{lcccccccccccccccc}
    \toprule
     & \multicolumn{4}{c}{\textit{Veach-Bidir}} & & \multicolumn{4}{c}{\textit{Veach-Ajar}} & & \multicolumn{4}{c}{\textit{Water-Caustic}} \\
    \cmidrule{2-5} \cmidrule{7-10} \cmidrule{12-15}
    Method & PSNR↑ & SSIM↑ & LPIPS↓ & Storage↓ & & PSNR↑ & SSIM↑ & LPIPS↓ & Storage↓ & & PSNR↑ & SSIM↑ & LPIPS↓ & Storage↓ \\
    \midrule
    Path Tracer 10 spp
    & 16.01 & 0.134 & 0.778 & --
    & & 9.276 & 0.098 & 0.651 & --
    & & 12.09 & 0.221 & 0.663 & -- \\

    Path Tracer 50 spp
    & 19.90 & 0.224 & 0.716 & --
    & & \colorbox{orange!25}{15.21} & 0.242 & 0.531 & --
    & & 12.86 & 0.319 & 0.573 & -- \\

    Particle Tracer 64 spp
    & \colorbox{orange!25}{25.29} & \colorbox{orange!25}{0.587} & \colorbox{orange!25}{0.365} & --
    & & 14.19 & 0.240 & 0.750 & --
    & & 12.31 & 0.542 & 0.476 & -- \\

    SPPM - 3 it
    & 18.01 & 0.146 & 0.696 & \colorbox{orange!25}{26.49}
    & & 14.58 & \colorbox{orange!25}{0.314} & \colorbox{orange!25}{0.516} & \colorbox{orange!25}{405.3}
    & & \colorbox{orange!25}{26.53} & \colorbox{orange!25}{0.664} & \colorbox{orange!25}{0.413} & \colorbox{orange!25}{14.16} \\

    EM-GMM
    & 13.52 & 0.095 & 0.823 & --
    & & 8.14 & 0.072 & 0.708 & --
    & & 12.39 & 0.185 & 0.702 & -- \\

    RenderFormer
    & 14.23 & 0.108 & 0.812 & --
    & & 8.56 & 0.081 & 0.692 & --
    & & 11.34 & 0.198 & 0.695 & -- \\

    GPF (Ours)
    & \colorbox{red!25}{37.61} & \colorbox{red!25}{0.926} & \colorbox{red!25}{0.338} & \colorbox{red!25}{18.33}
    & & \colorbox{red!25}{25.84} & \colorbox{red!25}{0.691} & \colorbox{red!25}{0.324} & \colorbox{red!25}{267.2}
    & & \colorbox{red!25}{27.12} & \colorbox{red!25}{0.936} & \colorbox{red!25}{0.158} & \colorbox{red!25}{9.724} \\

    \midrule
     & \multicolumn{4}{c}{\textit{Water-Caustic 2}} & & \multicolumn{4}{c}{\textit{Cornell Box}} & & \multicolumn{4}{c}{\textbf{Mean}} \\
    \cmidrule{2-5} \cmidrule{7-10} \cmidrule{12-15}
    Method & PSNR↑ & SSIM↑ & LPIPS↓ & Storage↓ & & PSNR↑ & SSIM↑ & LPIPS↓ & Storage↓ & & PSNR↑ & SSIM↑ & LPIPS↓ & Storage↓ \\
    \midrule
    Path Tracer 10 spp
    & 11.88 & 0.248 & 0.637 & --
    & & 30.49 & 0.654 & 0.497 & --
    & & 15.95 & 0.271 & 0.645 & -- \\

    Path Tracer 50 spp
    & 12.24 & 0.304 & 0.579 & --
    & & \colorbox{orange!25}{34.59} & \colorbox{orange!25}{0.850} & 0.403 & --
    & & 18.96 & 0.388 & 0.560 & -- \\

    Particle Tracer 64 spp
    & 12.08 & 0.528 & 0.489 & --
    & & 33.89 & 0.820 & \colorbox{orange!25}{0.370} & --
    & & 19.55 & \colorbox{orange!25}{0.543} & 0.490 & -- \\

    SPPM - 3 it
    & \colorbox{orange!25}{25.64} & \colorbox{orange!25}{0.632} & \colorbox{orange!25}{0.412} & \colorbox{orange!25}{16.26}
    & & 30.47 & 0.754 & 0.399 & \colorbox{orange!25}{21.63}
    & & \colorbox{orange!25}{23.05} & 0.502 & \colorbox{orange!25}{0.487} & \colorbox{orange!25}{96.77} \\

    EM-GMM
    & 10.47 & 0.192 & 0.688 & --
    & & 13.26 & 0.205 & 0.558 & --
    & & 11.56 & 0.150 & 0.696 & -- \\

    RenderFormer
    & 10.92 & 0.210 & 0.672 & --
    & & 32.14 & 0.738 & 0.452 & --
    & & 15.44 & 0.267 & 0.665 & -- \\

    GPF (Ours)
    & \colorbox{red!25}{28.72} & \colorbox{red!25}{0.962} & \colorbox{red!25}{0.196} & \colorbox{red!25}{12.34}
    & & \colorbox{red!25}{39.73} & \colorbox{red!25}{0.936} & \colorbox{red!25}{0.183} & \colorbox{red!25}{15.46}
    & & \colorbox{red!25}{31.80} & \colorbox{red!25}{0.890} & \colorbox{red!25}{0.240} & \colorbox{red!25}{64.61} \\

    \bottomrule
    \end{tabular}%
    }
    % \vspace{-5pt}
\end{table*}

\begin{figure*}[t]
    \centering
    \includegraphics[width=\textwidth]{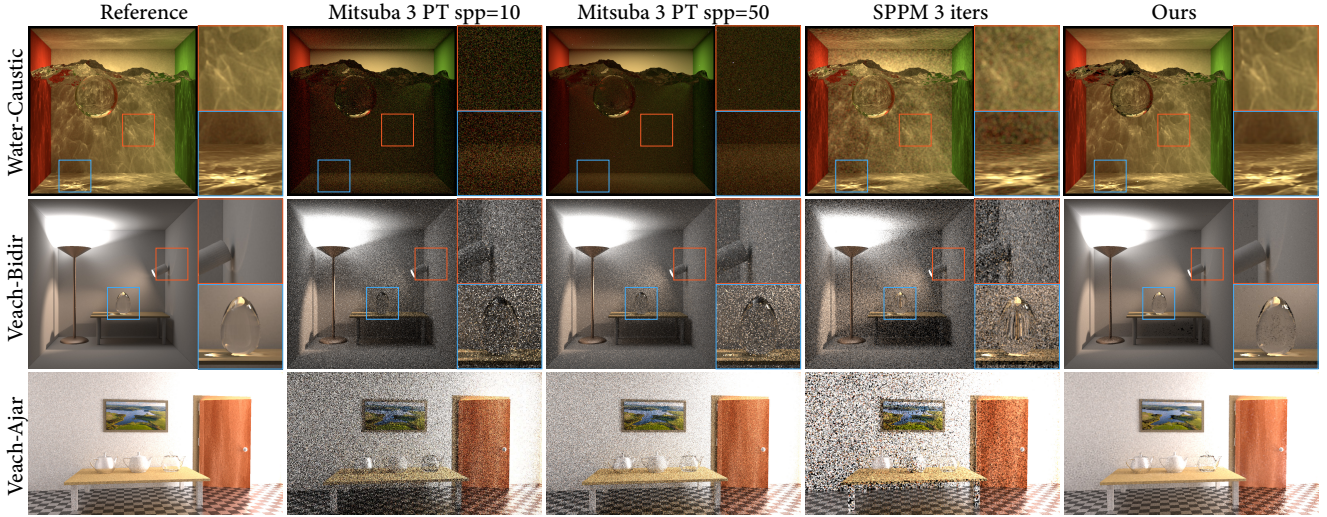}
    \vspace{-1em}
    \caption{
        \textbf{Qualitative comparison results.}
        We compare GPF with path tracing and \textit{SPPM} on scenes with complex indirect illumination and caustics.
Our method yields cleaner results with fewer noise artifacts and more accurate light transport, especially in challenging caustic and multi-bounce diffuse regions.
        Please \faSearchPlus~zoom in for details.
    }
    \label{fig:maincompare}
    \vspace{-.6em}
\end{figure*}

The results show that GPF achieves the highest overall performance across all metrics (PSNR~\cite{psnr}, SSIM~\cite{ssim}, LPIPS~\cite{lpips}).
In scenes dominated by indirect lighting (\textit{Veach-Bidir}, \textit{Veach-Ajar}), our method significantly surpasses all baselines in both SSIM and LPIPS, demonstrating its ability to capture smooth, high-frequency radiance variations while maintaining global consistency.
For caustic-dominated scenes (\textit{Water-Caustic}, \textit{Water–Caustic~2}), GPF achieves the best PSNR and substantially outperforms others in SSIM and LPIPS, highlighting its capacity to accurately model concentrated photon energy distributions through the learned continuous field.
While PSNR is a pixel-wise metric sensitive to small local deviations, our method preserves perceptual and structural fidelity much better, as reflected by large gains in SSIM and LPIPS.
Even on the simple \textit{Cornell Box}, GPF consistently outperforms all baselines across metrics, confirming its robustness and general applicability to diverse light transport phenomena. 

We further compare against two methods that represent alternative
paradigms for accelerating radiance computation, but both fail on
our scenes.
EM-GMM~\cite{jakob2011progressive}, the only prior method applying Gaussian mixtures to photon-based light transport, serves
as the most directly relevant baseline. However, its volumetric
kernel normalization ($\int G(\mathbf{x})\,dV = 1$) does not
hold for photons deposited on surfaces, resulting in incorrect
radiance estimates (see Sec.~\ref{sec:related_caching} for a
detailed discussion).
RenderFormer~\cite{zeng2025renderformer} represents the
feed-forward neural rendering paradigm, sharing GPF's
motivation of replacing repeated computation with a learned
mapping.
However, it is trained exclusively on opaque materials
and cannot handle transparency or refraction,
precluding caustics.
These failures underscore that complex light transport demands
physics-grounded representations rather than purely
statistical or data-driven alternatives.

\begin{table}[t]
    \centering
    \footnotesize
    \setlength{\tabcolsep}{4pt}
    \caption{\textbf{Comparison with Instant-NGP, 3DGS, AnySplat} on caustic scenes, with varying training view numbers.
    % We compare our method against neural rendering approaches trained with different numbers of views.
    \colorbox{red!25}{Best} and \colorbox{orange!25}{\nth{2} best} results are highlighted.}
    \vspace{-5pt}
    \label{tab:nerf_3dgs_comparison}
    \begin{tabular}{lccc|ccc}
    \toprule
     & \multicolumn{3}{c}{\textit{Water-Caustic}} & \multicolumn{3}{c}{\textit{Water-Caustic 2}} \\
    \cmidrule(lr){2-4} \cmidrule(lr){5-7}
    Method & PSNR↑ & SSIM↑ & LPIPS↓ & PSNR↑ & SSIM↑ & LPIPS↓ \\
    \midrule
    Instant-NGP~\cite{mueller2022instant} - 3 views
    & 13.42 & 0.466 & 0.748
    & 13.09 & 0.460 & 0.734 \\

    Instant-NGP~\cite{mueller2022instant} - 10 views
    & 12.59 & 0.473 & 0.618
    & 12.44 & 0.480 & 0.620 \\

    3DGS~\cite{3dgs} - 3 views
    & 16.76 & 0.508 & 0.493
    & 17.80 & 0.550 & 0.485 \\

    3DGS~\cite{3dgs} - 10 views
    & 26.58 & 0.812 & 0.246
    & 26.44 & 0.843 & 0.244 \\

    AnySplat~\cite{jiang2025anysplat} - 3 views
    & 17.24 & 0.522 & 0.476
    & 18.35 & 0.572 & 0.468 \\

    AnySplat~\cite{jiang2025anysplat} - 10 views
    & \colorbox{orange!25}{26.83} & \colorbox{orange!25}{0.825} & \colorbox{orange!25}{0.238}
    & \colorbox{orange!25}{26.71} & \colorbox{orange!25}{0.858} & \colorbox{orange!25}{0.235} \\

    \midrule
    GPF (Ours) - 3 views
    & \colorbox{red!25}{27.12} & \colorbox{red!25}{0.936} & \colorbox{red!25}{0.158}
    & \colorbox{red!25}{28.72} & \colorbox{red!25}{0.962} & \colorbox{red!25}{0.196} \\
    
    \bottomrule
    \end{tabular}
    % \vspace{-5pt}
\end{table}
\begin{figure*}
    \centering
    {\setlength{\tabcolsep}{1pt}
    \begin{tabular}{@{}c@{\hspace{1pt}}c@{\hspace{1pt}}c@{\hspace{1pt}}c@{\hspace{1pt}}c@{}}
    \small NeRF (3 views) &
    \small NeRF (10 views) &
    \small 3DGS (3 views) &
    \small 3DGS (10 views) &
    \small Ours (3 views)
    \end{tabular}}\\[2pt]
    \animategraphics[controls,buttonsize=.7em,autoplay,loop,palindrome,poster=16,width=1.\textwidth]{24}{Images/multiview-animation-water-caustic/composite/}{0}{23}
    \\
    {\small To view the animation, please use compatible software (\eg, \textit{Adobe Acrobat} or \textit{KDE Okular}).}
    \vspace{-.8em}
    \caption{\textbf{Multi-view comparison on Water-Caustic 2 with NeRF and 3DGS.}
    Unlike NeRF and 3DGS, which exhibit view-dependent artifacts under sparse supervision, Gaussian Photon Field (GPF) preserves consistent caustics and global illumination even with 3 views.}
    \label{fig:multi-view-animated-water-caustic}
  \vspace{-8pt}
\end{figure*}

\noindent\textbf{Comparison with Neural Scene Representations.}
To contextualize GPF within the broader paradigm of continuous scene representations, 
we compare it with neural radiance field methods such as Instant-NGP~\cite{mueller2022instant}, 
3DGS~\cite{3dgs}, and AnySplat~\cite{jiang2025anysplat} in Table~\ref{tab:nerf_3dgs_comparison}. We emphasize that this comparison is not intended as an equal-input benchmark. 
Neural baselines operate under fundamentally different assumptions: 
they rely solely on RGB images and camera poses, and primarily optimize for 
appearance reconstruction. In contrast, GPF assumes access to scene geometry, 
materials, and lighting, and explicitly models physically grounded light transport.
Our goal in this comparison is conceptual rather than competitive: 
to evaluate whether purely appearance-driven representations can faithfully 
reconstruct complex global illumination phenomena, \eg, caustics and 
specular–diffuse transport. Even when given significantly more input views than GPF, these methods struggle to recover high-frequency and physically-consistent illumination effects under 
sparse supervision. This suggests that physics-grounded modeling remains 
essential for accurate and view-consistent reconstruction of complex light transport.

\begin{table}[t]
\centering
\small
\setlength{\tabcolsep}{5pt}
% \vspace{-10pt}
\caption{\textbf{Ablation study on different components.}}
    \vspace{-5pt}
\label{tab:ablation_with_components}
\begin{tabular}{ccc|ccc}
    \toprule
    $w_{\text{init}}$ & $w_{\text{radius}}$ & $w_{\text{knn}}$ & PSNR↑ & SSIM↑ & LPIPS↓ \\
    \midrule
     & \checkmark&  & 26.73 & 0.8962 & 0.2152 \\
     & & \checkmark & 26.42 & 0.8847 & 0.2118 \\
     & \checkmark& \checkmark & 26.90 & 0.9026 & 0.2208 \\
    \checkmark &\checkmark  &  & 26.93 & 0.9128 & 0.1792 \\
    \checkmark &  &\checkmark  & 26.97 & 0.9118 & 0.1769 \\
    \checkmark & \checkmark & \checkmark & \textbf{27.12} & \textbf{0.9355} & \textbf{0.1579} \\
    \bottomrule
\end{tabular}
% \vspace{-7pt}
\end{table}
% \subsection{Ablation Study}
% \label{sec:exp_ablation}
\noindent\textbf{Ablation Studies.}
% \noindent\textbf{Ablation on Different Components.}
% Table~\ref{tab:ablation_with_components} analyzes the influence of key components in our framework, including
% the physically-grounded initialization ($w_{\text{init}}$),
% the radius-based neighborhood weighting ($w_{\text{radius}}$),
% and the $k$-nearest-neighbor fallback ($w_{\text{knn}}$).
% Without initialization, we randomly sample Gaussian primitives as the starting representation.
% Due to the lack of physically grounded photon priors, the optimization becomes unstable
% and tends to converge to suboptimal local minima, leading to a noticeable performance drop.
% When only $w_{\text{radius}}$ is used, regions with sparse photon coverage fail to query valid Gaussians,
% resulting in incomplete radiance reconstruction and reduced rendering quality.
% Conversely, using only $w_{\text{knn}}$ neglects the spatially adaptive weighting of photon density,
% causing over-smoothed illumination and weaker caustic details.
% Combining all three components yields the best overall performance,
% as $w_{\text{init}}$ provides a stable physical prior,
% $w_{\text{radius}}$ ensures spatial adaptivity,
% and $w_{\text{knn}}$ guarantees robust fallback in sparse regions.
Table~\ref{tab:ablation_with_components} analyzes the influence of three key components in our framework: the physically grounded initialization ($w_{\text{init}}$), the radius-based neighborhood weighting ($w_{\text{radius}}$), and the $k$-nearest-neighbor fallback ($w_{\text{knn}}$). Without initialization, Gaussian primitives are randomly sampled, and the lack of photon-based priors makes optimization unstable, often leading to suboptimal local minima. Using only $w_{\text{radius}}$ leaves sparse regions without valid Gaussians, resulting in incomplete radiance reconstruction. Using only $w_{\text{knn}}$ removes spatial adaptivity, producing over-smoothed illumination and weaker caustics. Combining all three components yields the best performance: $w_{\text{init}}$ provides a stable physical prior, $w_{\text{radius}}$ ensures spatial adaptivity, and $w_{\text{knn}}$ offers a robust fallback in sparse regions. Further ablation studies are provided in the appendix.

\subsection{Qualitative Evaluation}
\label{sec:exp_qualitative}
% Figure~\ref{fig:maincompare} presents qualitative comparisons across four representative scenes 
% (\textit{Veach--Bidir}, \textit{Veach--Ajar}, \textit{Water--Caustic}, and \textit{Water--Caustic~2}); 
% results on the \textit{Cornell Box} are provided in the supplementary material.
Figure~\ref{fig:maincompare} presents qualitative comparisons across three representative scenes, with more exhaustive results in supplementary material.
% For the \textit{Water-Caustic} scenes, GPF produces highly accurate and visually sharp caustics, 
% faithfully capturing the subtle light patterns refracted through the water surface. 
% In contrast, \textit{SPPM - 3 iters} yields noticeably blurred illumination due to the limited photon budget per iteration, 
% and \textit{Mitsuba~3 Path Tracer} fails to reconstruct detailed caustic structures even at 50~spp.  
% For the \textit{Veach--Bidir} scene, which involves complex multi-bounce diffuse transport and glossy interreflections, 
% GPF achieves a clean, noise-free reconstruction with well-defined highlights and sharp caustic details, 
% particularly around the transparent egg and wall reflections.  
% Both \textit{SPPM - 3 iters} and \textit{Path Tracer} exhibit significant noise and structural artifacts.  
% Similarly, in the \textit{Veach--Ajar} scene, GPF delivers a smooth and stable illumination field with accurate indirect lighting, 
% whereas other integrators suffer from high-frequency noise and incomplete light propagation.
% These results demonstrate that GPF effectively unifies the physical accuracy of photon-based rendering 
% with the smoothness and stability of a continuous learned radiance field, 
% achieving visually superior results under both caustic and indirect illumination conditions.
For the \textit{Water-Caustic} scene, GPF produces sharp, accurate caustics that capture the subtle refractive patterns, while \textit{SPPM - 3 iters} yields blurred illumination due to its limited photon budget and path tracing~\cite{Mitsuba3} (PT) fails to recover detailed structures even at 50~spp. In the \textit{Veach--Bidir} scene, which features complex multi-bounce diffuse transport and glossy interreflections, GPF achieves a clean, noise-free reconstruction with well-defined highlights, unlike \textit{SPPM - 3 iters} and path tracing, which exhibit noise and structural artifacts. Similarly, in the \textit{Veach-Ajar} scene, GPF provides smooth, stable illumination with accurate indirect lighting, whereas other integrators suffer from high-frequency noise and incomplete light transport. Overall, GPF combines the physical accuracy of photon-based rendering with the smoothness of a continuous learned radiance field, delivering superior results for both caustic and indirect illumination.

% Furthermore, as illustrated in Figure~\ref{fig:multi-view-animated-water-caustic}, our Gaussian Photon Field (GPF) maintains strong consistency across multiple novel viewpoints. 
% Despite large changes in camera pose and viewing direction, both direct and indirect lighting remain coherent, with stable caustic structures and diffuse shading. 
% This demonstrates the robustness and view-reusability of our learned radiance field, effectively amortizing photon-based light transport across views. 
% Comprehensive per-view visualizations for all tested scenes are provided in the supplementary material.
Furthermore, as shown in Figure~\ref{fig:multi-view-animated-water-caustic}, our photon field enables strong consistency across novel viewpoints. Despite large changes in camera pose and viewing direction, both direct and indirect lighting remain coherent, with stable caustics and diffuse shading. This highlights the robustness and view-reusability of our learned field, effectively amortizing photon-based light transport across views. Additional visualizations for all scenes are provided in the appendix.

\begin{table}[t]
    % \vspace{-10pt}
    \centering
    \footnotesize
    \setlength{\tabcolsep}{4pt}
    \caption{\textbf{Rendering time (in seconds) comparison}, across five scenes.}
    \vspace{-5pt}
    \label{tab:rendering_time}
    \begin{tabular}{lrrrrr}
    \toprule
    Method & \textit{V-Bidir} &\textit{ V-Ajar} & \textit{W-Caustic} & \textit{W-Caustic 2} & \textit{Cornell} \\
    \midrule
    PT 10 spp    & 3.9 & 3.8 & 3.7 & 3.9& 2.5 \\
    PT 50 spp    & 4.3 & 5.0 & 7.4 & 7.5 & 4.5 \\
    SPPM - 3 iters    & 23.4 & 42.0 & 26.4 & 24.7 & 32.8 \\
    SPPM - 1000 iters\;\; & 7,793.3 & 14,010.0 & 8,790.0 & 8246.7 & 10,923.0 \\
    GPF (Ours)  & 6.3 & 30.1 & 16.34 & 15.3 & 26.7 \\
    \bottomrule
    \end{tabular}
    \vspace{-10pt}
\end{table}
%

% \noindent\textbf{Efficiency Analysis.}
\subsection{Efficiency Analysis}
\textbf{Rendering Time.} Table~\ref{tab:rendering_time} compares the rendering time per image across all scenes.
While the \textit{Path Tracer} (spp=10 and spp=50) provides fast but noisy results, and \textit{SPPM - 3 iters} yields moderate accuracy at the cost of higher computation,
our method achieves comparable visual quality  to the \textit{SPPM - 1000 iters} %
 \begin{wrapfigure}{r}{0.4\textwidth}
    \centering
    \vspace{-5pt}
    \includegraphics[width=0.38\textwidth]{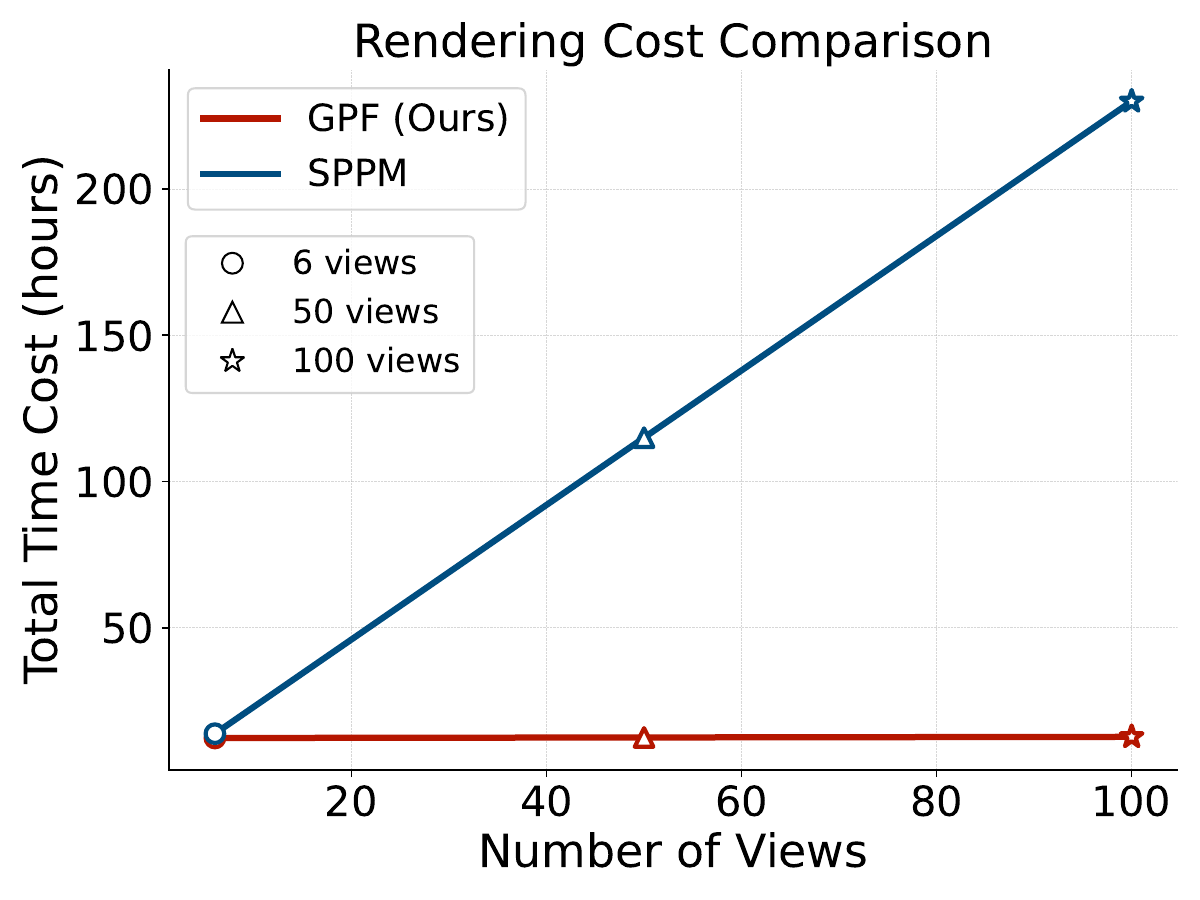}
    \vspace{-7pt}
    \caption{Total rendering cost comparison. GPF amortizes a one-time training cost, becoming more efficient than SPPM beyond 6 views.}
    \label{fig:cost_comparison_wrap}
    % \vspace{-40pt}
\end{wrapfigure}
reference while being orders of magnitude faster.
Specifically, \textit{SPPM - 1000 iters}—used only for generating the reference and final radiance ground truth—requires several hours per frame (up to 14,000 seconds),
whereas GPF produces results of similar fidelity within only 6–30 seconds per image.
This efficiency stems from reusing the learned radiance field across views, eliminating repeated photon tracing and progressive iterations.

\noindent\textbf{Total Computational Cost.}
% In terms of training cost, 
GPF requires a one-time preprocessing cost: generating ground-truth radiance 
maps (2.3h $\times$ 3 views = 6.9h) and training the field (5.4h), totaling 
approximately 12.3h. After training, rendering a novel view takes only 15s, compared to 2.3h for \textit{SPPM - 1000 iters}. Figure~\ref{fig:cost_comparison_wrap} shows
the total time as a function of the view number. Even accounting for the full preprocessing cost (GT generation + training), 
GPF becomes more efficient than SPPM after \textbf{6 views}, 
achieving 1.9$\times$ speedup at 10 views and \textbf{18$\times$} at 100 views. 
This demonstrates clear practical value for multi-view rendering scenarios 
such as simulation, synthetic data generation, and virtual production.

% \input{Tables/ablation_gaussian_num}
% \input{Tables/ablation_knn}
% \input{Tables/all_ablations}

% \noindent\textbf{Ablation on Different Number of Gaussian Primitives.}
% Table~\ref{tab:ablation_combined} (top right) investigates the effect of varying the number of Gaussian primitives. 
% Increasing the number of Gaussians consistently improves image quality, as reflected by higher PSNR/SSIM and lower LPIPS, owing to denser photon coverage and finer radiance approximation. 
% However, beyond 100K primitives, the performance gain becomes marginal while rendering time increases notably. 
% We therefore adopt 100K Gaussians as a balanced configuration between quality and efficiency.

% \noindent\textbf{Ablation on Different k-NN Parameter.}
% Table~\ref{tab:ablation_combined} (bottom right) analyzes the influence of the $k$-nearest-neighbor parameter on radiance aggregation. 
% A too small $k$ (\eg, $k{=}1$) leads to unstable estimation and higher noise due to insufficient photon support, 
% while an excessively large $k$ oversmooths high-frequency lighting and caustic details. 
% The overall performance remains stable for $k$ between 3 and 5, showing nearly identical image quality and computation time. 
% We adopt $k{=}3$ as our default setting for its robustness and simplicity.
\section{Conclusion}
We introduce the Gaussian Photon Field (GPF), a physically-grounded radiance representation that unifies photon mapping and neural fields. By reformulating per-view kernel density estimation into a continuous Gaussian field, GPF enables efficient, view-reusable global illumination while preserving physical accuracy. Experiments show that GPF achieves photon-level accuracy with orders-of-magnitude faster rendering and strong multi-view consistency. This formulation bridges classical light transport simulation and learnable radiance representations, pointing toward physically accurate, differentiable rendering of dynamic and volumetric scenes. 
Future work will explore leveraging GPF’s efficiency and differentiability for inverse caustics.

% ---------------------------------------------------------------
% Acknowledgements
\section*{Acknowledgements}
We thank the anonymous reviewers for their insightful comments and suggestions.

% ---------------------------------------------------------------
% Bibliography
\bibliographystyle{splncs04}
\bibliography{main_ECCV}

% ---------------------------------------------------------------
% Supplementary Material (merged as appendix for arXiv)
\appendix
\beginsupplement

\section*{Supplementary Material}

In this supplementary material, we provide additional implementation details for reproducibility (Section \ref{sec:supp_implementation}), our complete ablation study (Section \ref{sec:supp_ablation}) and additional experimental results (Section \ref{sec:supp_results}). 
We also include a supplementary video illustrating the multi-view rendering consistency of our Gaussian Photon Field across diverse scenes.

\section{Implementation Details}
\label{sec:supp_implementation}

We implement both our baseline \textit{Stochastic Progressive Photon Mapping}~\cite{hachisuka2009sppm} and the proposed \textit{Gaussian Photon Field (GPF)} integrator within the \textit{Mitsuba~3}~\cite{Mitsuba3} rendering framework, built upon the \texttt{Dr.Jit} differentiable runtime~\cite{Jakob2022DrJit}. Both integrators share the same scene interface and BSDF/geometry infrastructure, allowing direct performance and quality comparison.

\subsection{SPPM Integrator}
\label{sec:impl_sppm}
Our baseline follows the standard formulation of stochastic progressive photon mapping~\cite{hachisuka2009sppm} (\ref{code:sppm}), consisting of alternating \emph{photon tracing} and \emph{camera ray tracing} stages. 
Each iteration emits $N_p$ photons from all light sources and stores them in a KD-tree for radiance estimation.
The photon radius $r_t$ is updated after each iteration according to
\begin{equation}
r_{t+1} = r_t \sqrt{\frac{t+\alpha}{t+1}},
\end{equation}
where $\alpha = 0.7$ controls the blending rate between old and new photon contributions.
For the experiments, we run $T = 1000$ iterations to produce the ground-truth reference radiance $L_{\mathrm{ref}}$.

\subsection{GPF Integrator}
\label{sec:impl_gpf}
The GPF integrator extends the SPPM pipeline by replacing the per-view kernel density estimation (KDE) stage with a continuous, learnable Gaussian Photon field.
We initialize a set of $M=N_p=100,000$ anisotropic Gaussian primitives from the first SPPM iteration's photon map, using the photon position as the mean $\boldsymbol{\mu}_i$, initial scale $\mathbf{s}_i=0.01$, and random orientation $\mathbf{q}_i$.
Each Gaussian stores the photon flux $\boldsymbol{\Phi}_i$ as its radiance contribution.

During training, camera rays are traced through the scene using a hybrid path tracing integrator (\ref{code:gpf-render}).
At each diffuse surface intersection $\mathbf{x}$, we query nearby Gaussians within a radius $r=0.02$ or $k=3$ nearest neighbors using a \emph{hybrid radius+kNN} search (\ref{code:gpf-query}). 

\begin{code}
\caption{
\normalsize{\textbf{Stochastic Progressive Photon Mapping (SPPM) baseline} has three stages: (1) photon tracing to build KD-tree, (2) camera pass with KDE-based radiance gathering at first diffuse hits, and (3) progressive radius reduction following $r_{t+1}=r_t\sqrt{\frac{t+\alpha}{t+1}}$. SPPM is used as a baseline and our ground truth generator (with $T=1000$ iterations).}}
\label{code:sppm}
\vspace{-0.5em}
\begin{pycode}
def sppm(scene, camera, light, T, Np, r0=0.02, alpha=0.7):
    L_accum, rt = 0.0, r0
    for t in range(T):
        # 1) Photon tracing: emit Np photons and build KD-tree
        P = PHOTON_TRACE(scene, light, Np)         # store (pos, flux, wi) on diffuse surfaces
        KDT = BUILD_KDTREE(P)

        # 2) Camera pass: first diffuse hit uses KDE (radius-(*@\textcolor{codegreen}{$r_t$}@*) gathering)
        L_frame = 0.0
        for pix in camera:
            ray = PRIMARY_RAY(camera, pix)
            hit, beta, prev_delta = TRACE_TO_FIRST_DIFFUSE(scene, ray)
            if hit is None:
                L_frame += 0.0; continue
            if prev_delta and IS_EMITTER(hit):
                L_frame += beta * EMITTER_RADIANCE(hit); continue
            N = RADIUS_QUERY(KDT, hit.x, rt)
            Lp = KDE_GATHER(N, hit) # radiance: (*@\textcolor{codegreen}{$L_p=\frac{1}{\pi r_t^2}\sum \Phi_j f_r$}@*)
            L_frame += beta * Lp

        # 3) Progressive radius update
        rt = r0 * ((t + alpha) / (t + 1))**0.5
        L_accum += L_frame
    return L_accum / T
\end{pycode}
\end{code}

% Each Gaussian contributes to the queried radiance as:
% \[
% L(\mathbf{x}) = 
% \frac{\sum_i \alpha_i\, w_i(\mathbf{x})\, \mathrm{SH}(\mathbf{h}_i, \omega_o)}
% {\sum_i w_i(\mathbf{x}) + \epsilon},
% \]
where $w_i$ is the anisotropic Gaussian weight and $\epsilon = 10^{-6}$ ensures stability.
Queries are implemented using a KD-tree structure built from Gaussian centers, rebuilt periodically as positions are optimized.

We only supervise \textbf{visible surface points} collected from $K$ camera views ($K=3$–30 depending on the scene).
Each visible point’s ground-truth radiance $L_{\mathrm{GT}}(\mathbf{x})$ is obtained from the reference SPPM after 1000 iterations. 
%Supervision is provided on \textbf{visible surface points}

We train the field by minimizing the MSE loss
\begin{equation}
\mathcal{L} = \frac{1}{N}\sum_{\mathbf{x}} \|L_{\mathrm{GPF}}(\mathbf{x}) - L_{\mathrm{ref}}(\mathbf{x})\|_2^2,
\end{equation}
where $N$ is the number of visible surface points,
using the Adam optimizer with a learning rate of $5\times10^{-4}$.
Training is performed for 10k iterations, each sampling a random subset of visible points from randomly selected views.
\begin{code}
\caption{
\label{code:gpf-render}
\normalsize{\textbf{Camera ray tracing with Gaussian queries}. 
At diffuse surfaces, we query the radiance from GPF, multiply by the BSDF, and terminate. Specular/glossy segments are traced explicitly.}}
\vspace{-0.3em}
\begin{pycode}
def render_with_gpf(scene, camera, G, r_query=0.02, k_min=3, kd_tree):
    """
    Differentiable camera pass that reuses the learned Gaussian field.
    Query occurs only at diffuse surfaces; specular/glossy bounces are traced.
    """
    L_img = zeros_like(camera.pixels)
    for pix in camera.pixels:
        ray = generate_camera_ray(camera, pix)
        L_pix, beta, prev_delta = 0.0, 1.0, True
        while True:
            hit = intersect(ray, scene)
            if not hit: break

            if is_emitter(hit) and prev_delta:
                L_pix += beta * emitter_radiance(hit, ray.dir)
                break

            wo, fs, pdf, bsdf_type = sample_bsdf(hit, return_type=True)

            if is_diffuse(bsdf_type):
                # Query learned radiance field and multiply by BSDF
                Li = query_gaussian_radiance(
                hit.position, wo, G, kd_tree,
                r_query, k_min)
                Lg = Li * fs  # Apply BSDF modulation
                L_pix += beta * Lg
                break  # terminate at first diffuse
            else:
                beta *= fs / pdf
                prev_delta = is_delta(bsdf_type)
                ray = spawn_ray(hit, wo)

        L_img[pix] = L_pix
    return L_img
\end{pycode}
\end{code}

All differentiable operations (Gaussian weighting and gradient accumulation) are implemented in \texttt{Dr.Jit}~\cite{Jakob2022DrJit} with full GPU vectorization. 

\begin{code}
\caption{
\label{code:gpf-query}
\normalsize{\textbf{Hybrid Gaussian radiance query}. We first use a radius query to preserve locality and supplement with kNN to avoid empty neighborhoods. A soft normalization by $\max(\sum w_i,\varepsilon)$ ensures stability.}}
\vspace{-0.3em}
\begin{pycode}
def query_gaussian_radiance(x, view_dir, G, kd_tree, r=0.02, k_min=3, eps=1e-6):
    """
    x:        surface point (query)
    view_dir: outgoing direction (to camera)
    G:        Gaussian params [(mu, sigma, quat, color), ...]
    kd_tree:  built over {mu}
    r:        radius for ball query (default: 0.02)
    k_min:    minimal neighbors guaranteed by kNN supplement (default: 3)
    returns:  radiance estimate at x
    """
    # Phase 1: radius query
    idx_r = kd_tree.ball_query(x, r)        
    # Phase 2: supplement with kNN if sparse
    if len(idx_r) < k_min:
        idx_knn = kd_tree.knn_query(x, k_min)
        idx = dedup(idx_r + idx_knn)       
    else:
        idx = idx_r
    # Phase 3: accumulate weighted contributions
    L = 0.0
    Z = 0.0
    for j in idx:
        mu, sigma, quat, color = G[j]
        R = quat_to_matrix(quat)                 # 3x3 rotation
        d  = x - mu
        dL = R.T @ d          # to local frame
        dn = (dL[0]/sigma[0], dL[1]/sigma[1], dL[2]/sigma[2])
        w_gauss = exp(-0.5 * dot(dn, dn))        # anisotropic kernel
        # soft distance falloff (smooth outside r)
        dist = norm(d)
        if dist <= r:
            w_dist = 1.0
        else:
            t = (dist - r) / max(r, 1e-6)
            w_dist = exp(-3.0 * t * t)
        w = w_gauss * w_dist
        L += color * w
        Z += w
    return L / max(Z, eps)   
\end{pycode}
\end{code}

\section{Ablation Study}
\label{sec:supp_ablation}

\begin{table}[t]
\centering
\small
\setlength{\tabcolsep}{5pt}
\caption{\textbf{Ablation study on the number of Gaussian primitives.}}
\label{tab:ablation_gaussian_num}
\begin{tabular}{c|ccc|c}
    \toprule
    \# Gaussians & PSNR↑ & SSIM↑ & LPIPS↓ & Time (s)↓ \\
    \midrule
    10K  & 24.32 & 0.8521 & 0.2847 & 7.230 \\
    50K  & 26.18 & 0.8892 & 0.2156 & 12.70 \\
    100K & 27.12 & 0.9355 & 0.1579 & 16.37 \\
    200K & 27.25 & 0.9403 & 0.1564 & 24.35 \\
    \bottomrule
\end{tabular}
\end{table}

\begin{table}[t]
\centering
\small
\setlength{\tabcolsep}{5pt}
\caption{\textbf{Ablation study on the k-NN parameter $k$.}}
\label{tab:ablation_knn}
\begin{tabular}{c|ccc|c}
    \toprule
    $k$ & PSNR↑ & SSIM↑ & LPIPS↓ & Time (s)↓ \\
    \midrule
    1  & 25.87 & 0.8756 & 0.2314 & 16.21 \\
    3  & 27.12 & 0.9355 & 0.1579 & 16.37 \\
    5  & 27.18 & 0.9372 & 0.1574 & 16.48 \\
    10 & 27.08 & 0.9342 & 0.1595 & 16.34 \\
    \bottomrule
\end{tabular}
\end{table}

\begin{figure*}[t]
    \centering
    \includegraphics[width=\textwidth]{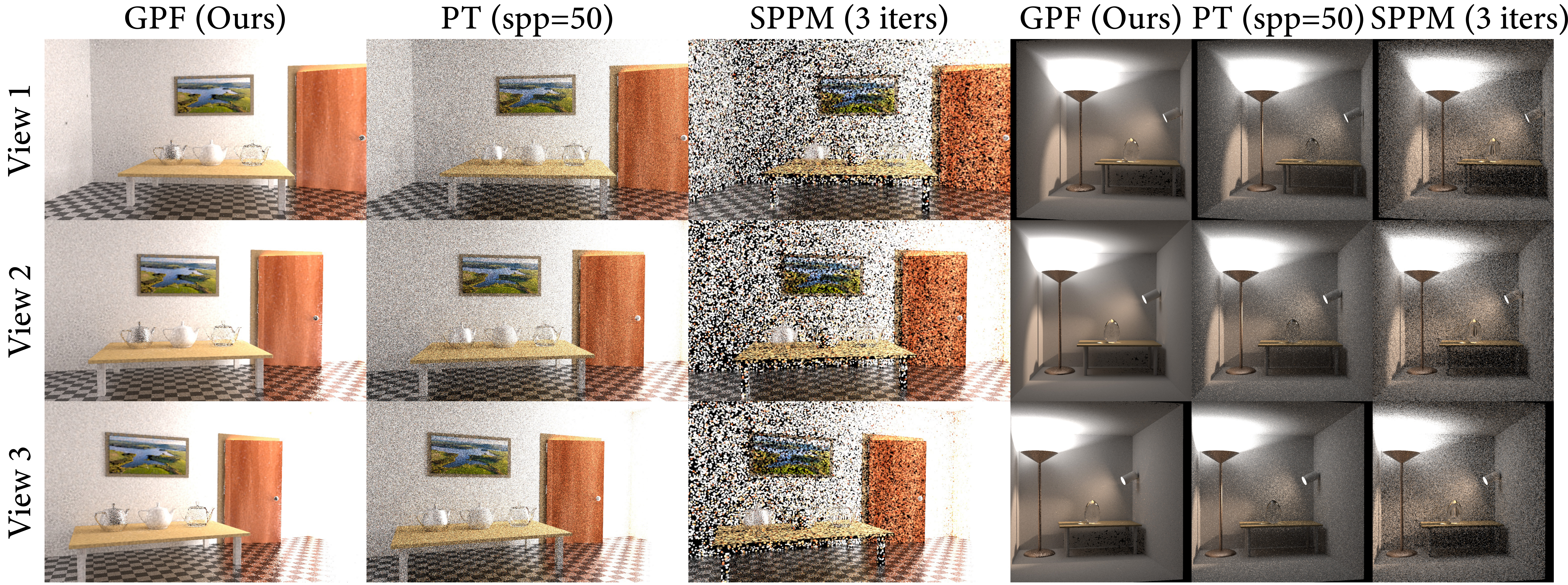}
    \caption{
        \textbf{Multi-view rendering results.}
        We compare our solution with \textit{Mitsuba~3 Path Tracer}~\cite{Mitsuba3}, \textit{SPPM}~\cite{hachisuka2009sppm}, \textit{3DGS}~\cite{3dgs} and \textit{Instant-NGP}~\cite{mueller2022instant} across water-caustic.
        Our method produces cleaner results with significantly fewer noise artifacts and more accurate light transport, especially in challenging caustic.
        Please \faSearchPlus~zoom in for more details.
    }
    \label{fig:multi3}
\end{figure*}

\noindent\textbf{Ablation on Different Number of Gaussian Primitives.}
Table~\ref{tab:ablation_gaussian_num} investigates the effect of varying the number of Gaussian primitives. 
Increasing the number of Gaussians consistently improves image quality, as reflected by higher PSNR/SSIM and lower LPIPS, owing to denser photon coverage and finer radiance approximation. 
However, beyond 100K primitives, the performance gain becomes marginal while rendering time increases notably. 
We therefore adopt 100K Gaussians as a balanced configuration between quality and efficiency.

\noindent\textbf{Ablation on Different k-NN Parameter.}
Table~\ref{tab:ablation_knn} analyzes the influence of the $k$-nearest-neighbor parameter on radiance aggregation. 
A too small $k$ (\eg, $k{=}1$) leads to unstable estimation and higher noise due to insufficient photon support, 
while an excessively large $k$ oversmooths high-frequency lighting and caustic details. 
The overall performance remains stable for $k$ between 3 and 5, showing nearly identical image quality and computation time. 
We adopt $k{=}3$ as our default setting for its robustness and simplicity.

\begin{figure*}[t]
    \centering
    \includegraphics[width=\textwidth]{Images/multi_view_water_1_5.pdf}
    \caption{
        \textbf{Multi-view rendering results.}
        We compare our solution with \textit{Mitsuba~3 Path Tracer}~\cite{Mitsuba3}, \textit{SPPM}~\cite{hachisuka2009sppm}, \textit{3DGS}~\cite{3dgs} across water-caustic.
        Our method produces cleaner results with significantly fewer noise artifacts and more accurate light transport, especially in challenging caustic.
        Please \faSearchPlus~zoom in for more details.
    }
    \label{fig:multi1}
\end{figure*}

\begin{figure*}[t]
    \centering
    \includegraphics[width=\textwidth]{Images/multi_view_water_2_4.pdf}
    \caption{
        \textbf{Multi-view rendering results.}
        We compare our solution with \textit{Mitsuba~3 Path Tracer}~\cite{Mitsuba3}, \textit{SPPM}~\cite{hachisuka2009sppm}, \textit{3DGS}~\cite{3dgs} and \textit{Instant-NGP}~\cite{mueller2022instant} across water-caustic.
        Our method produces cleaner results with significantly fewer noise artifacts and more accurate light transport, especially in challenging caustic.
        Please \faSearchPlus~zoom in for more details.
    }
    \label{fig:multi2}
\end{figure*}

\section{Additional Experiment Results}
\label{sec:supp_results}

We also provide a supplementary video demonstrating the multi-view rendering consistency across all scenes.

\noindent\textbf{Multi-View Rendering Results on Veach-Ajar and Veach-Bidir}. 
We present multi-view rendering results on the \textit{Veach-Ajar} and \textit{Veach-Bidir} scenes from three viewpoints (Figure~\ref{fig:multi3}). 
Our method is compared with Path Tracing (50~spp) and SPPM (3 iterations). 
The results demonstrate that our approach produces cleaner images with reduced noise and more stable indirect illumination across views compared to both baselines.

\noindent\textbf{Multi-View Rendering Results on Water-Caustic}. 
We present multi-view rendering results on the \textit{Water-Caustic} (Figure~\ref{fig:multi1}) scene using four viewpoints. 
We compare our method with 3DGS under both \textit{3-view} and \textit{10-view} supervision, as well as with Path Tracing (50~spp) and SPPM (3 iterations). 
The results indicate that our method achieves sharper caustic structures, reduced noise levels, and more stable multi-view radiance behavior compared to the baseline approaches.

\noindent\textbf{Multi-View Rendering Results on Water-Caustic 2}. We present multi-view rendering results on the \textit{Water-Caustic 2} scene (Figure~\ref{fig:multi2}), using four viewpoints. We compare our method with 3DGS and Instant-NGP under both \textit{3-view} and \textit{10-view} supervision, as well as with Path Tracing (50~spp) and SPPM (3 iterations). The results show that our approach produces cleaner caustics, fewer noise artifacts, and more consistent light transport across views compared to all baselines.

\end{document}